%% file: sn-article.tex
\documentclass[pdflatex,sn-mathphys-num]{sn-jnl}% Math and Physical Sciences Numbered Reference Style
\usepackage{graphicx}%
\usepackage{multirow}%
\usepackage{amsmath,amssymb,amsfonts}%
\usepackage{amsthm}%
\usepackage{mathrsfs}%
\usepackage[title]{appendix}%
\usepackage{xcolor}%
\usepackage{textcomp}%
\usepackage{manyfoot}%
\usepackage{booktabs}%
\usepackage{algorithm}%
\usepackage{algorithmicx}%
\usepackage{algpseudocode}%
\usepackage{listings}%

\usepackage{graphicx}%
\usepackage{multirow}%
\usepackage{amsmath,amssymb,amsfonts}%
\usepackage{amsthm}%
\usepackage{mathrsfs}%
\usepackage[title]{appendix}%
\usepackage[table,dvipsnames]{xcolor}%
\usepackage{textcomp}%
\usepackage{manyfoot}%
\usepackage{booktabs}%
\usepackage{algorithm}%
\usepackage{algorithmicx}%
\usepackage{algpseudocode}%
\usepackage{listings}%

\usepackage[shortlabels]{enumitem}
\usepackage{array}
\usepackage{setspace}
\usepackage{caption}
\usepackage{tabularx}
\usepackage{makecell}
\usepackage{nicematrix}
\usepackage{hyperref}

\definecolor{replay}{RGB}{255,192,0}
\definecolor{reg}{RGB}{239,76,162}
\definecolor{pea}{RGB}{18,116,100}
\definecolor{zsd}{RGB}{68, 84, 106}
\definecolor{transfer}{RGB}{68, 114, 196}
\definecolor{last}{RGB}{166,107,211}
\definecolor{avg}{RGB}{112, 173, 71}
\definecolor{forget}{RGB}{39,129,114}
\definecolor{bwt}{RGB}{15,51,45}
\definecolor{runi}{RGB}{251,229,214}
\definecolor{amul}{RGB}{250,230,231}
\definecolor{native}{RGB}{217,243,239}
\hypersetup{
 linkcolor = last,
 citecolor = avg,
 colorlinks = true,
 urlcolor = transfer
}
\usepackage[table,dvipsnames]{xcolor}%
\theoremstyle{thmstyleone}%
\theoremstyle{thmstyletwo}%

\theoremstyle{thmstylethree}%

\begin{document}

\title[Article Title]{Continual Learning for VLMs: A Survey and Taxonomy Beyond Forgetting}


%\author*[1,2]{\fnm{First} \sur{Author}}\email{iauthor@gmail.com}



%\affil*[1]{\orgdiv{Department}, \orgname{Organization}, \orgaddress{\street{Street}, \city{City}, \postcode{100190}, \state{State}, \country{Country}}}

%\affil[2]{\orgdiv{Department}, \orgname{Organization}, \orgaddress{\street{Street}, \city{City}, \postcode{10587}, \state{State}, \country{Country}}}

%\affil[3]{\orgdiv{Department}, \orgname{Organization}, \orgaddress{\street{Street}, \city{City}, \postcode{610101}, \state{State}, \country{Country}}}

 \title[Continual Learning for VLMs]{Continual Learning for VLMs: A Survey and Taxonomy Beyond Forgetting}

\author*[1]{\fnm{Yuyang} \sur{Liu}}\email{liuyuyang13@pku.edu.cn}
\author[1]{\fnm{Qiuhe} \sur{Hong}}
\author[2]{\fnm{Linlan} \sur{Huang}}
\author[3]{\fnm{Alexandra} \sur{Gomez-Villa}}
\author[3]{\fnm{Dipam} \sur{Goswami}}
\author[1]{\fnm{Tiantian} \sur{Peng}}
\author[2]{\fnm{Xialei} \sur{Liu}}
\author[3]{\fnm{Joost} \spfx{van de} \sur{Weijer}}
\author*[1]{\fnm{Yonghong} \sur{Tian}}\email{yhtian@pku.edu.cn}

\affil[1]{\orgdiv{Shenzhen Graduate School}, \orgname{Peking University}, \orgaddress{\city{Shenzhen}, \country{China}}}

\affil[2]{\orgdiv{VCIP, TMCC, College of Computer Science}, \orgname{Nankai University}, \orgaddress{\city{Tianjin}, \country{China}}}

\affil[3]{\orgname{Computer Vision Center, Universitat Autònoma de Barcelona}, \orgaddress{\city{Barcelona}, \country{Spain}}}


\abstract{Vision-language models (VLMs) and Multimodal Large Language Models (MLLMs) have significantly advanced artificial intelligence through robust cross-modal alignment and zero-shot generalization.
However, enabling them to learn continually from non-stationary data remains a major challenge, as their cross-modal alignment and generalization capabilities are particularly vulnerable to catastrophic forgetting.
Unlike traditional unimodal continual learning (CL), VLMs face unique challenges such as cross-modal feature drift, parameter interference due to shared architectures, and zero-shot capability erosion.
Furthermore, generative MLLMs exhibit a unique “alignment tax,” while catastrophic forgetting extends beyond factual amnesia to a systemic collapse of deep Chain-of-Thought (CoT) reasoning.
This survey presents the first comprehensive diagnostic review bridging continual learning for predictive VLMs including generative MLLMs. We analyze these failure modes and introduce a challenge-driven taxonomy: (1) \textit{Multi-Modal Replay Strategies} addressing explicit and implicit memory drift; (2) \textit{Cross-Modal Regularization} enforcing topological and geometric alignment; (3) \textit{Parameter-Efficient Adaptation} utilizing dynamic routing and subspace projections. We review the evolution of evaluation protocols, focusing on the emergence of dual-track benchmarks for Domain and Ability CL and fine-grained diagnostic evaluations of Chain-of-Thought (CoT) reasoning. We conclude by outlining future research directions, with an emphasis on compositional zero-shot learning, embodied AI with sensor fusion, and autonomous agentic ecosystems. All resources are available at: \url{https://github.com/YuyangSunshine/Awesome-Continual-learning-of-Vision-Language-Models}.}

\keywords{Vision-language models (VLMs), Continual Learning, Multimodal Large Language Models (MLLMs), Survey}



\maketitle

\input{introduction}

\input{background}

\input{challenges}

\input{methods}

\input{methods_mllms}

\input{Evaluation}

\input{experiments}

\input{future_work}


\newpage
\bibliography{sn-bibliography}

\end{document}

%% file: introduction.tex
\section{Introduction}\label{sec:intro}
Vision-Language Models (VLMs) have emerged as a dominant paradigm for multimodal learning by acquiring cross-modal representations from web-scale image--text corpora. Their architectural landscape ranges from early dual-encoder models, such as CLIP~\cite{radford2021learning}, ALIGN~\cite{jia2021scaling}, and BLIP~\cite{li2022blip}, to recent Multimodal Large Language Models (MLLMs), which integrate visual perception with autoregressive language models, including LLaVA~\cite{liu2023visual} and Qwen-VL~\cite{bai2023qwen}.
While dual-encoder VLMs demonstrate strong zero-shot performance in cross-modal retrieval and classification, MLLMs substantially broaden their applicability through instruction-following and open-ended multimodal reasoning. These models have achieved impressive results across a wide range of tasks, including image--text retrieval~\cite{rong2026retrieval,xie2023ra}, visual question answering (VQA)~\cite{antol2015vqa}, and multimodal reasoning~\cite{yue2024mmmu}. Despite their architectural differences, both paradigms are built upon the pre-train-and-transfer framework, where large-scale pre-training establishes a shared multimodal representation that generalizes effectively to downstream tasks~\cite{radford2021learning}.

Despite these advances, existing VLMs are typically developed under the assumption of a static training distribution. In real-world applications, however, embodied agents, scientific robots, and personalized assistants continuously encounter evolving environments in which new data arrives sequentially and previously unseen concepts emerge over time. This mismatch raises a fundamental challenge:

\textit{How can VLMs continuously acquire new knowledge while preserving previously learned capabilities?}

Naively fine-tuning them on new tasks inevitably results in \textit{catastrophic forgetting}~\cite{mccloskey1989catastrophic}. The continual learning (CL) community has proposed a wide range of techniques, such as regularization~\cite{kirkpatrick2017overcoming}, rehearsal~\cite{rebuffi2017icarl}, and dynamic network expansion~\cite{DEN}, to mitigate forgetting in unimodal models~\cite{parisi2019continual,delange2021continual}. 
However, extending continual learning to multimodal foundation models presents a substantially different set of challenges due to their architectural characteristics and scale. Early studies on multimodal continual learning (MMCL)~\cite{yu2026recent, wei2024promptmm} mainly focused on relatively compact models and constrained benchmarks, such as audio--visual classification on AVE~\cite{tian2018ave}, where multimodal architectures remain comparatively homogeneous. In contrast, modern vision-language foundation models typically combine frozen vision encoders, lightweight cross-modal projectors, and large language models, creating new challenges for continual adaptation while preserving the representations acquired during large-scale pre-training.

\begin{figure}[!t]  
	\centering   
	\includegraphics[width = 0.85\linewidth]{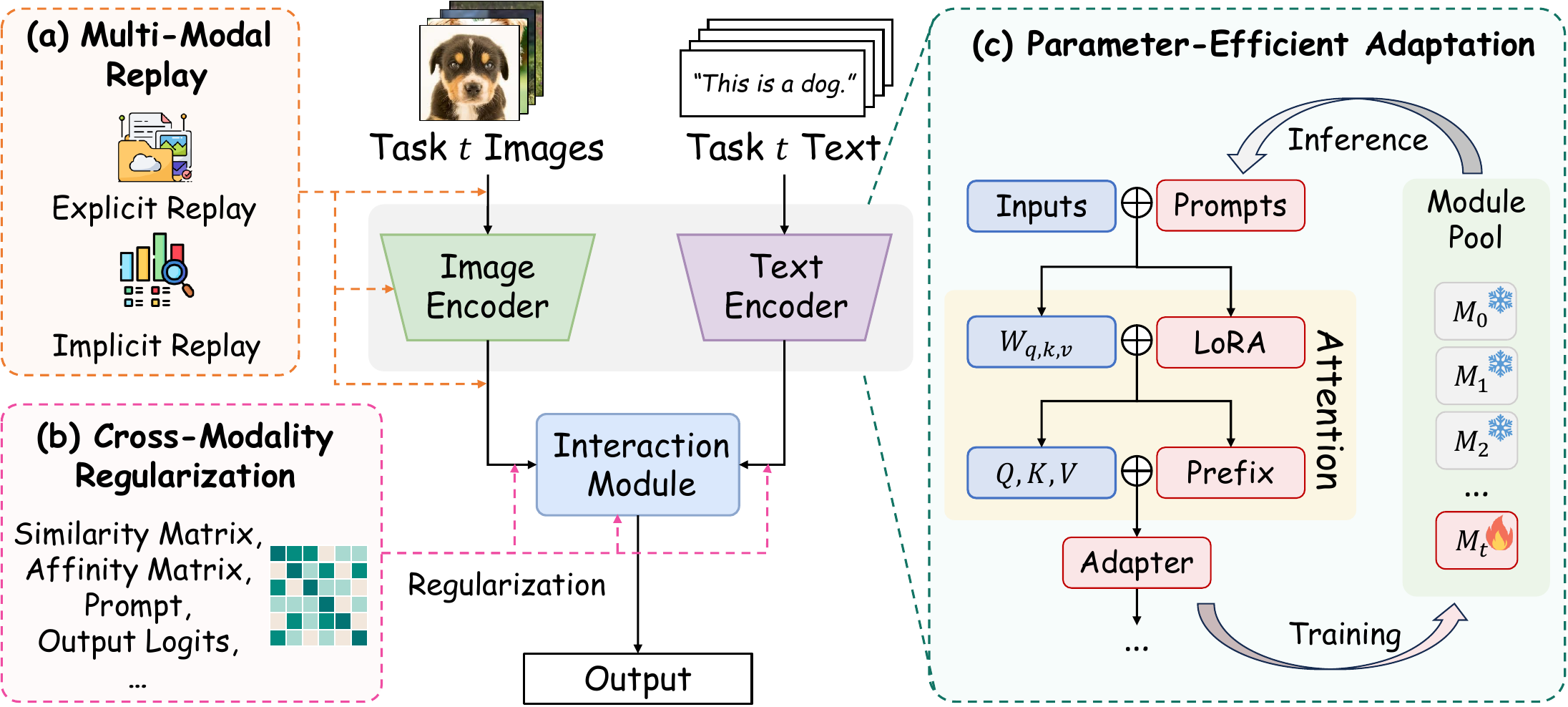}     
	\caption{Taxonomy of Continual Learning Strategies for Vision-Language Models. Existing methods are grouped into three main paradigms: (a) \textbf{Multi-Modal Replay}, which preserves previous knowledge through data rehearsal; (b) \textbf{Cross-Modal Regularization}, which constrains model updates to mitigate forgetting; and (c) \textbf{Parameter-Efficient Adaptation}, which retains task-specific knowledge through parameter isolation and efficient adaptation.} 
        \captionsetup{justification=centering}
	\label{fig:cls_sum}  
    \vspace{-10pt}
\end{figure}

Although recent surveys have reviewed the development of MMCL~\cite{yu2026recent}, they provide limited discussion of the challenges that arise in foundation-scale vision-language models. Throughout this survey, we organize these challenges into three representative categories (see Section~\ref{sec:chanllenge}). First, Cross-Modal Feature Drift refers to the progressive distortion of the shared embedding space during sequential adaptation~\cite{yan2022generative, huang2025MGCLIP}. Second, Shared Module Interference describes the unintended modification of parameters in shared components responsible for cross-modal interaction~\cite{hu2022lora, yu2024boosting}. Third, Zero-Shot Capability Erosion captures the degradation of open-vocabulary representations caused by continual updates, leading to reduced zero-shot generalization~\cite{wu2025synthetic, peng2025gnspgradientnullspace}. Critically, as models scale to generative MLLMs, this erosion uniquely manifests as a systemic collapse of Chain-of-Thought (CoT) logic~\cite{gao2026many, guo2026mllmctbench}. Furthermore, beyond these shared vulnerabilities, the autoregressive nature and instruction-following alignment of MLLMs introduce an additional layer of specific pathologies, such as the ``alignment tax'' and modality imbalance, which we exclusively dissect in Section~\ref{sec:method_mllms}.

Despite a surging interest in deploying foundation models for complex, real-world continuous scenarios , such as generalist medical diagnostics~\cite{chen2026forging}, continual GUI learning~\cite{yao2026cgl}, or long-term instruction following~\cite{marouf2025no}, the field lacks a dedicated survey that systematically deconstructs these micro-level failure mechanisms. To bridge this critical gap, this paper presents the first comprehensive, diagnostic review of continual learning spanning both predictive VLMs and generative MLLMs. 

To provide a unified perspective on the evolution from vision-language models (VLMs) to multimodal large language models (MLLMs) under continual learning scenarios, we organize existing approaches according to the challenges they address and the mechanisms they adopt. As summarized in Figure~\ref{fig:cls_sum}, existing methods can be broadly grouped into three complementary paradigms. The main contributions of this survey are summarized as follows:
\begin{itemize}
    \item We provide a systematic analysis of the challenges in continual learning for VLMs and MLLMs, including representation drift, cross-modal alignment degradation, zero-shot capability erosion, and emerging issues in generative MLLMs such as instruction alignment degradation and reasoning capability preservation;

    \item We establish a unified taxonomy of continual vision-language learning methods from three perspectives: \textit{multimodal replay}, \textit{cross-modal regularization}, and \textit{parameter-efficient adaptation}. Representative approaches, including replay strategies, structural constraints, prompt tuning, low-rank adaptation, and dynamic routing mechanisms, are comprehensively reviewed and compared;

    \item We summarize the evolution of evaluation protocols and benchmarks, highlighting the transition from conventional continual learning settings toward more comprehensive evaluations for vision-language foundation models, including domain adaptation, zero-shot generalization, and capability-oriented continual learning;

    \item We discuss remaining challenges and future directions, including scalable long-horizon continual adaptation, multimodal reasoning preservation, benchmark standardization, and the deployment of continual vision-language models in real-world scenarios.
\end{itemize}

The remainder of this survey is organized as follows. Section~\ref{sec:back} introduces the necessary background. Section~\ref{sec:chanllenge} discusses the key challenges and underlying mechanisms in continual vision-language learning, which motivate the taxonomy presented in Section~\ref{sec:method}. Section~\ref{sec:method_mllms} reviews continual learning methods for MLLMs. Section~\ref{sec:evaluation} summarizes existing evaluation protocols and recent dual-track benchmarks. Finally, Section~\ref{sec:discussion} discusses open challenges and future research directions.

%% file: background.tex
\section{Preliminaries and Fundamental Concepts}\label{sec:back}

This section introduces the background necessary for the remainder of the survey. We first establish the notation used throughout the paper, followed by a brief overview of continual learning (CL) and the main architectures of Vision-Language Models (VLMs). We then introduce two representative paradigms in continual vision-language learning: \textit{Continual Fine-tuning (CFT)} and \textit{Continual Pre-training (CPT)}. For reference, the primary mathematical notations used throughout the paper are summarized in Table~\ref{tab:notation}.

\input{table/notation}

\subsection{Traditional Continual Learning}

Continual learning (CL) aims to enable models to continuously absorb new knowledge from a series of sequential tasks while avoiding catastrophic forgetting~\cite{mccloskey1989catastrophic,delange2021continual}. To address this challenge, traditional CL methods primarily focus on solving the ``stability-plasticity dilemma''~\cite{parisi2019continual,lifelong_machine}: the model must stably retain old knowledge (stability) while flexibly learning new knowledge (plasticity). Based on different strategies for balancing this dilemma, these methods can be roughly divided into the following categories:

\paragraph{Regularization-based Methods}
Regularization-based methods \cite{kirkpatrick2017overcoming, zenke2017continual, li2017learning, liu2021l3doc} mitigate catastrophic forgetting by constraining updates to parameters that are important for previously learned tasks. A representative example is Elastic Weight Consolidation (EWC)~\cite{kirkpatrick2017overcoming}, which estimates parameter importance using the Fisher information matrix $\mathbf{F}$. The objective is formulated as
\begin{equation}
    \mathcal{L}(\theta) = \mathcal{L}_{\text{new}}(\theta) + \frac{\lambda}{2} \sum_i \mathbf{F}_i (\theta_i - \theta_i^*)^2,
\end{equation}
where $\theta_i^*$ represents the parameters after completing previous tasks and $\lambda$ is a balancing coefficient. Other methods such as Synaptic Intelligence (SI)~\cite{zenke2017continual} and Memory Aware Synapses (MAS)~\cite{aljundi2018memory} similarly quantify parameter importance by accumulating parameter changes and gradient information.

\paragraph{Dynamic Architecture Methods} These methods~\cite{DEN,rusu2016progressive} dynamically adjust the model's architecture to accommodate new tasks, typically by isolating task-specific parameters. For example, Progressive Neural Networks (PNNs)~\cite{rusu2016progressive} freeze the main network after learning a task and add a new set of columns (sub-networks) for each new task, with lateral connections to leverage features from previous tasks. This approach completely avoids forgetting but leads to a model size that grows linearly with the number of tasks. 

\paragraph{Rehearsal Methods} These methods~\cite{rebuffi2017icarl,recall,MC,RBN,liu2023augmented} mitigate forgetting by storing a small subset of data from past tasks in a memory buffer. During training on a new task, these stored samples are replayed alongside the new data to refresh the model's memory of past tasks. Methods~\cite{aljundi2018memory} like iCaRL~\cite{rebuffi2017icarl} combine rehearsal with a nearest-mean-of-exemplars classifier for class-incremental learning, while others~\cite{DGR} focus on generating pseudo-samples to avoid privacy and storage issues associated with raw data replay.

\subsection{Vision-Language Models}

Vision-Language Models (VLMs)~\cite{zhang2024vision} aim to jointly model visual and textual information for cross-modal representation learning and understanding. Existing VLMs~\cite{du2022survey} can be broadly divided into two categories: dual-encoder architectures and fusion-based architectures.

Dual-encoder models, exemplified by CLIP~\cite{radford2021learning}, employ separate encoders for images and text and align their representations in a shared embedding space through contrastive learning. Specifically, linear projection layers map visual and textual features into a common embedding space, where alignment is optimized using contrastive objectives such as InfoNCE~\cite{oord2018representation}:
\begin{equation}
    \mathcal{L}_{\text{InfoNCE}} =
-\log
\frac{\exp(\text{sim}(\mathbf{f}_V,\mathbf{f}_T)/\tau)}
{\sum_j \exp(\text{sim}(\mathbf{f}_i,\mathbf{f}_j)/\tau)},
\end{equation}
where $\mathbf{f}_V$ and $\mathbf{f}_T$ denote the visual and textual feature representations, respectively, $\text{sim}(\cdot,\cdot)$ is a similarity function, and $\tau$ is a temperature parameter. Other representative dual-encoder models, including SigLIP~\cite{zhai2023sigmoid} and BLIP~\cite{li2022blip}, follow the same design principle.

Although most discriminative tasks primarily use only text embeddings of CLIP for zero-shot evaluation~\cite{radford2021learning} using cosine similarities, recent works~\cite{udandarao2023sus,li2025closing,goswami2026cross} proposed the fusion or mixing of both vision and text embeddings for improved classification and retrieval tasks. Beyond inter-modal tasks, CLIP has also been used for intra-modal classification and retrieval tasks (image-image or text-text) in~\cite{kordopatis2025ilias,mistretta2025cross,magistri2026isoclip} or in text-to-image generation~\cite{gal2023image,ruiz2023dreambooth}.
% IsoCLIP~\cite{magistri2026isoclip} decomposed CLIP projectors to identify inter- and intra-modal alignment operators which could be exploited for improved intra-modal performance.

Fusion-based architectures, such as ALBEF~\cite{li2021align}, perform deeper cross-modal interaction by integrating visual and textual representations after initial encoding. These models commonly employ cross-attention mechanisms to capture fine-grained relationships between modalities. A typical cross-attention operation is formulated as
\begin{equation}
    \mathbf{F} = \text{softmax}\left(\frac{(\mathbf{W_Q}\mathbf{S}_2)(\mathbf{W_K}\mathbf{S}_1)^\top}{\sqrt{d}}\right)\mathbf{W_V}\mathbf{S}_1,
\end{equation}
where $\mathbf{S}_1$ and $\mathbf{S}_2$ denote feature sequences from two modalities, $d$ represents the query/key dimension, and $\mathbf{W_Q}$, $\mathbf{W_K}$, and $\mathbf{W_V}$ are learnable projection matrices. In contrast to dual-encoder VLMs, where each modality is encoded independently, cross-attention uses $\mathbf{S}_2$ as queries and $\mathbf{S}_1$ as keys and values, enabling token-level interactions across modalities.

In VLM-CL, the choice of backbone architecture is closely related to the target task:
\begin{itemize}
    \item \textbf{Discriminative tasks} (\textit{e.g.,} classification and retrieval) commonly adopt dual-encoder models such as CLIP~\cite{radford2021learning}, benefiting from their strong zero-shot capability and modular architecture, which facilitate continual learning.
    \item \textbf{Generative tasks} (\textit{e.g.,} VQA) have traditionally relied on fusion-based architectures, including ViLT~\cite{kim2021vilt}, ALBEF~\cite{li2021align}, and FLAVA~\cite{singh2022flava}. Recent generative VLMs have shifted toward MLLM architectures, where visual representations are projected into the embedding space of pre-trained LLMs. Representative examples include MiniGPT-4~\cite{zhu2023minigpt} and LLaVA~\cite{liu2023visual}, which adopt lightweight projection modules~\cite{li2023blip2}; these architectures have also been explored in continual learning settings such as GMM~\cite{cao2024generative} and HiDe-LLaVA~\cite{guo2025hide}.
    \item \textbf{Fine-grained spatial reasoning tasks} (\textit{e.g.,} continual object detection) require explicit alignment between linguistic concepts and visual regions. For example, TAM~\cite{zhang2024learning} adopts GLIP~\cite{li2021glip} as its backbone to align region-level visual features with language representations, improving open-vocabulary generalization in continual learning scenarios.
\end{itemize}

A key challenge for any VLM architecture in continual learning is maintaining the pre-trained cross-modal alignment. Continual updates can easily disrupt this alignment and erode zero-shot generalization, a problem particularly acute in fusion-based models.

\subsection{Parameter-Efficient Continual Learning}

Continual learning with large pre-trained models (PTMs)~\citep{mcdonnell2023ranpac, wang2023isolation, gao2023unified, marouf2024weighted, zhou2025revisiting} increasingly relies on parameter-efficient strategies to incorporate new knowledge while preserving previously acquired capabilities. Existing approaches can be broadly categorized into two groups: parameter-efficient fine-tuning (PEFT)~\cite{he2021towards} and prompt learning~\cite{wei2021finetuned}.

PEFT methods, including Adapter~\cite{houlsby2019parameter}, LoRA~\cite{hu2022lora}, and Prefix-tuning~\cite{li2021prefix}, adapt pre-trained models by updating only a small number of additional or task-specific parameters. A representative direction within PEFT is to reduce interference between tasks by enforcing orthogonality among task-specific parameters. For example, O-LoRA~\cite{wang2023orthogonal} constrains the low-rank adaptation parameters of new tasks to lie in an orthogonal subspace with respect to previous tasks. Specifically, the orthogonality between the adaptation matrices $\mathbf{A}_t$ and $\mathbf{A}_i$ for two different tasks $t$ and $i$ is formulated as:
\begin{equation}
    \mathbf{A}^\mathsf{T}_t\mathbf{A}_i = 0, \quad \forall t \neq i.
\end{equation}

N-LoRA~\cite{yang2025parameter} further enhances this strategy by improving orthogonality constraints and reducing parameter interference. Similarly, PEGP~\cite{qiao2025gradient} applies orthogonal gradient projection based on Singular Value Decomposition (SVD) of task data, preventing new updates from disrupting previously learned feature representations.

The second paradigm, prompt learning, introduces task-specific prompts at the input level while keeping the pre-trained model fixed. Methods such as L2P~\cite{wang2022learning} and DualPrompt~\cite{wang2022dualprompt} learn task-specific prompt tokens or vectors, enabling knowledge adaptation without modifying the backbone parameters and thereby reducing catastrophic forgetting. These parameter-efficient approaches are also important for MLLM in continual learning, where full-model fine-tuning is often computationally prohibitive and may compromise the generalization ability of pre-trained multimodal models.

\subsection{Continual Fine-tuning vs. Continual Pre-training}

VLM-CL can be studied from two complementary perspectives, which differ in their objectives, data regimes, and adaptation strategies.

\paragraph{Continual Fine-tuning (CFT)}
CFT is the prevailing setting in current VLM-CL research. It assumes a fixed pre-trained VLM backbone (e.g., CLIP~\cite{radford2021learning} or LLaVA~\cite{liu2023visual}) and aims to adapt the model to a sequence of downstream tasks involving new domains, classes, or task formats. For MLLMs, CFT is often instantiated as continual instruction tuning (CIT), where new instruction datasets, domains, or capabilities are introduced sequentially. To preserve knowledge acquired during large-scale pre-training, parameter updates are typically limited to a small subset of parameters through PEFT techniques, such as prompts, adapters, and LoRA. The central challenge of CFT is to achieve an effective balance between plasticity for acquiring new knowledge and stability for retaining previous task-specific knowledge and general zero-shot capabilities. In practice, CFT is usually conducted on relatively small-scale and well-curated datasets.

\paragraph{Continual Pre-training (CPT)}
CPT aims to enable foundation models to continuously incorporate new knowledge from evolving multimodal data streams, representing a more general setting for lifelong multimodal learning. However, research on CPT remains relatively limited. Existing efforts investigate topics such as importance-aware continual pre-training (TIC-CLIP~\cite{garg2023tic}), preservation of cross-task compatibility and representation topology during sequential image-text training (CTP~\cite{zhu2023ctp}), and the use of regularization, PEFT, and model merging strategies at the pre-training scale (FoMo-in-Flux~\cite{roth2024practitioner}).

Most existing VLM-CL studies focus on the CFT setting due to its lower computational cost and closer connection to practical downstream applications. Accordingly, this survey primarily reviews CFT approaches while discussing CPT as an important future direction toward developing foundation models capable of adapting to continuously evolving world knowledge.

%% file: table/notation.tex
\begin{table}[t]
\centering
\small
\caption{Summary of notations used throughout the paper.}
\label{tab:notation}
\vspace{-5pt}
\begin{tabular}{ll}
\toprule
\textbf{Symbol} & \textbf{Description} \\
\midrule
\multicolumn{2}{l}{\textit{\textbf{General Continual Learning (CL) Notation}}} \\
\midrule
$T$ & Total number of sequential tasks. \\
$t$ & Index of the current task. \\
$p_i^t$ & Performance on task $t$ after training up to task $i$. \\
$\theta$ & The set of model parameters. \\
$\theta_i^*$ & Optimal parameters after training on task $i$. \\
$\mathcal{L}(\theta)$ & Overall loss function of the model. \\
$\lambda$ & A balancing coefficient for regularization. \\
$\mathbf{F}_i$ & The Fisher Information Matrix for task $i$. \\
\midrule
\multicolumn{2}{l}{\textit{\textbf{Vision-Language Model (VLM) Notation}}} \\
\midrule
$f_V, f_T$ & Feature embeddings for vision and text. \\
$\tau$ & Temperature parameter in contrastive loss. \\
$sim(\cdot, \cdot)$ & Similarity function. \\
$\mathbf{Q, K, V}$ & Query, Key, and Value matrices in attention. \\
$\mathbf{W}, \Delta \mathbf{W}$ & Weight matrix and its LoRA update. \\
$\mathbf{A}, \mathbf{B}$ & Low-rank decomposition matrices for LoRA. \\
$r$ & Rank of low-rank matrices. \\
$P$ & A learnable prompt. \\
\bottomrule
\end{tabular}
\vspace{-5pt}
\end{table}

%% file: challenges.tex
\begin{figure}[!t]  
	\centering   
	\includegraphics[width = 0.7\linewidth]{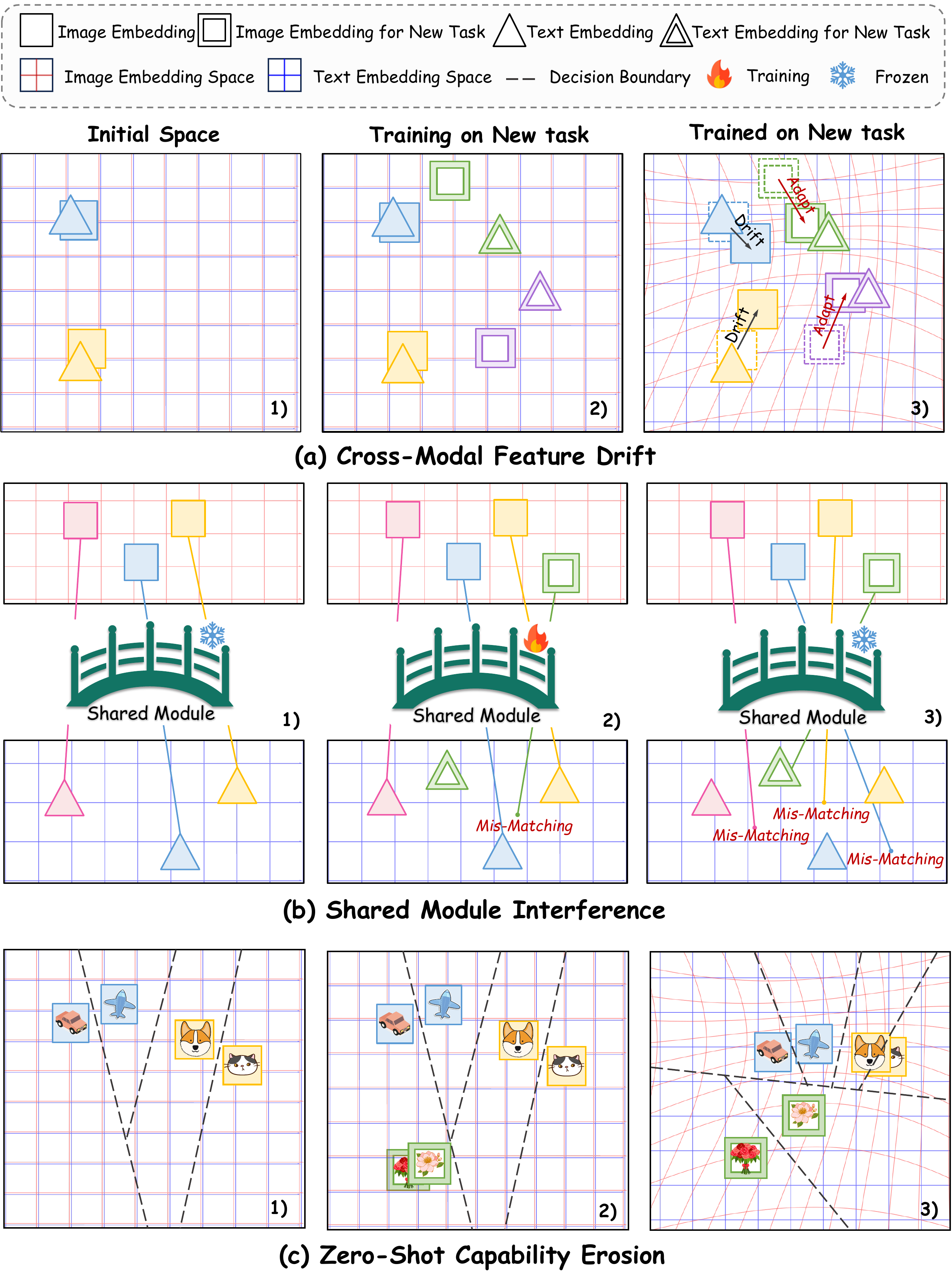}  
	\caption{Illustration of three core challenges in VLM-CL. \textbf{(a) Cross-Modal Feature Drift:} Initially aligned image-text representations become misaligned after adaptation to new tasks, weakening cross-modal correspondence for previously learned concepts. \textbf{(b) Shared Module Interference:} Updates for new tasks overwrite shared module parameters, causing performance degradation on both previous and current tasks. \textbf{(c) Zero-Shot Capability Erosion:} Fine-tuning on narrow tasks distorts the original representation space, leading to semantic collapse of zero-shot concepts and reduced generalization ability.} %   \captionsetup{justification=centering}
	\label{fig:challenges} 
    \vspace{-10pt}
\end{figure}

\section{Core Challenges in VLM-CL}~\label{sec:chanllenge}

Although VLM-based continual learning inherits fundamental challenges from traditional continual learning, such as the stability--plasticity trade-off~\cite{delange2021continual,masana2022class}, its multimodal nature introduces additional challenges that do not arise in unimodal settings. In particular, catastrophic forgetting in VLM-CL can be characterized from three closely related perspectives, as illustrated in Figure~\ref{fig:challenges}.

\subsection{Cross-Modal Feature Drift}

VLMs rely on the alignment between visual and textual representations established during large-scale pre-training~\cite{radford2021learning,li2022blip,xie2023ra}. However, this alignment can be disrupted during continual adaptation. As model parameters are updated over sequential tasks, the visual and textual embedding spaces may gradually become misaligned, resulting in \textit{Cross-Modal Feature Drift}~\cite{wang2021continual,huang2024class}. As illustrated in Figure~\ref{fig:challenges} (a), image--text representations corresponding to previously learned concepts may shift away from their original correspondence. Such misalignment~\cite{yu2025language,huang2024class,jha2024clap4clip,huang2025MGCLIP} weakens the association between visual and linguistic modalities and can degrade downstream capabilities such as multimodal retrieval. Existing approaches commonly address this issue through cross-modal regularization (Section~\ref{sec:method}).

\subsection{Shared Module Interference}

Fusion-based VLMs, such as ALBEF~\cite{li2021align} and ViLT~\cite{kim2021vilt}, rely on shared components, including cross-attention modules, to facilitate interactions between modalities. During continual learning, these shared components can become vulnerable to interference from sequential updates. As shown in Figure~\ref{fig:challenges} (b), optimization for new tasks may modify parameters that are important for previously learned tasks, reducing the effectiveness of cross-modal fusion. This interference can lead to performance degradation on earlier tasks and highlights the challenge of adapting shared architectures without compromising existing knowledge.

\subsection{Zero-Shot Capability Erosion}

A major advantage of VLMs such as CLIP~\cite{radford2021learning} is their ability to generalize to unseen concepts through a semantically structured embedding space. However, continual fine-tuning may gradually compromise this capability, leading to \textit{Zero-Shot Capability Erosion}~\cite{wortsman2022robust,zheng2023preventing,huang2025MGCLIP}. As models adapt to sequential tasks, their representations may become increasingly specialized toward recently observed data. Figure~\ref{fig:challenges} (c) illustrates this effect, where fine-tuning on a specific task (e.g., flower classification) alters the original embedding structure and reduces the separability of previously distinguishable concepts (e.g., ``dog'' and ``cat''). This phenomenon reflects the inherent trade-off between acquiring new knowledge and maintaining broad zero-shot generalization in lifelong multimodal learning.

%% file: methods.tex
\begin{figure}[!t]     
	\centering 
	\includegraphics[width = 1\linewidth]{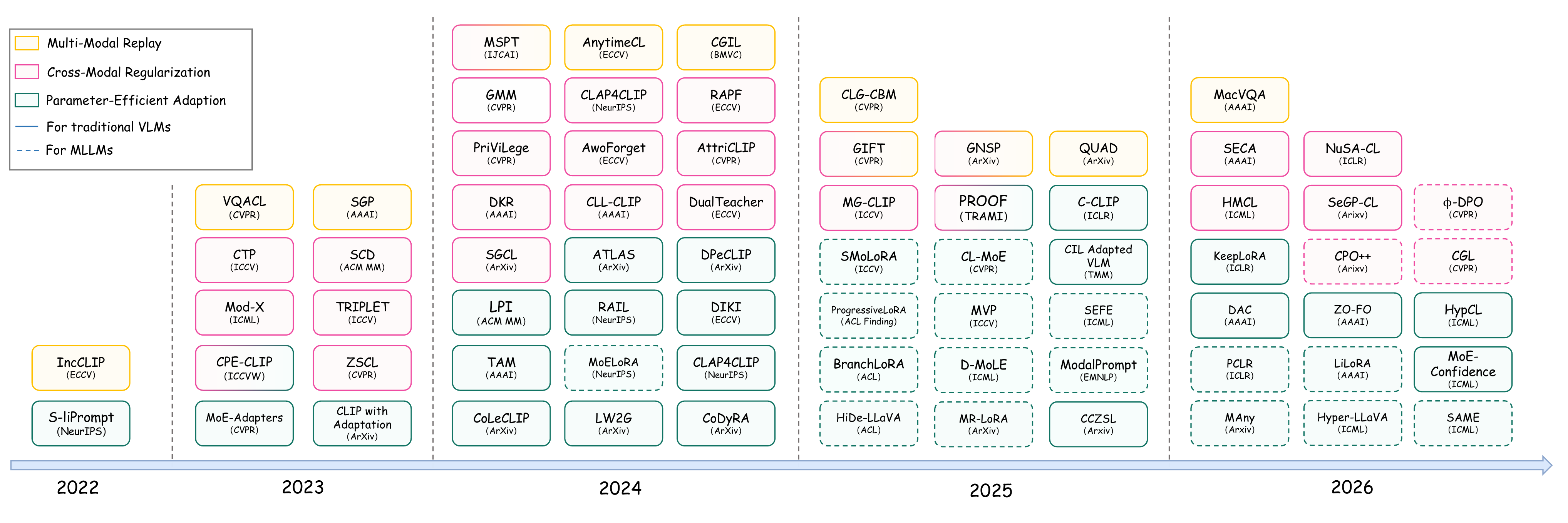}  
    \vspace{-15pt}
	\caption{Summary of VLM continual learning methods published in recent years. Several methods combine approaches; we categorize them according to their main contribution.
    } 
        \captionsetup{justification=centering}
	\label{fig:cls-method} 
    \vspace{-5pt}
\end{figure}

\section{Taxonomy of VLM-CL Methods}\label{sec:method}

As vision-language models (VLMs) are increasingly deployed in open-world scenarios, they face unique challenges in continual learning (CL). This section establishes a taxonomy of existing approaches, categorizing them into three dominant paradigms: (A) \textbf{Multi-Modal Replay Strategies}, (B) \textbf{Cross-Modal Regularization}, and (C) \textbf{Parameter-Efficient Adaptation}, as shown in Figure~\ref{fig:cls-method}. It is worth noting that these three categories of methods are not completely independent. They each focus on solving one or more core challenges and often appear in a hybrid form.

\subsection{Multi-Modal Replay Strategy (MMRE)}

\begin{figure}[!t] 
	\centering   
	\includegraphics[width = 0.8\linewidth]{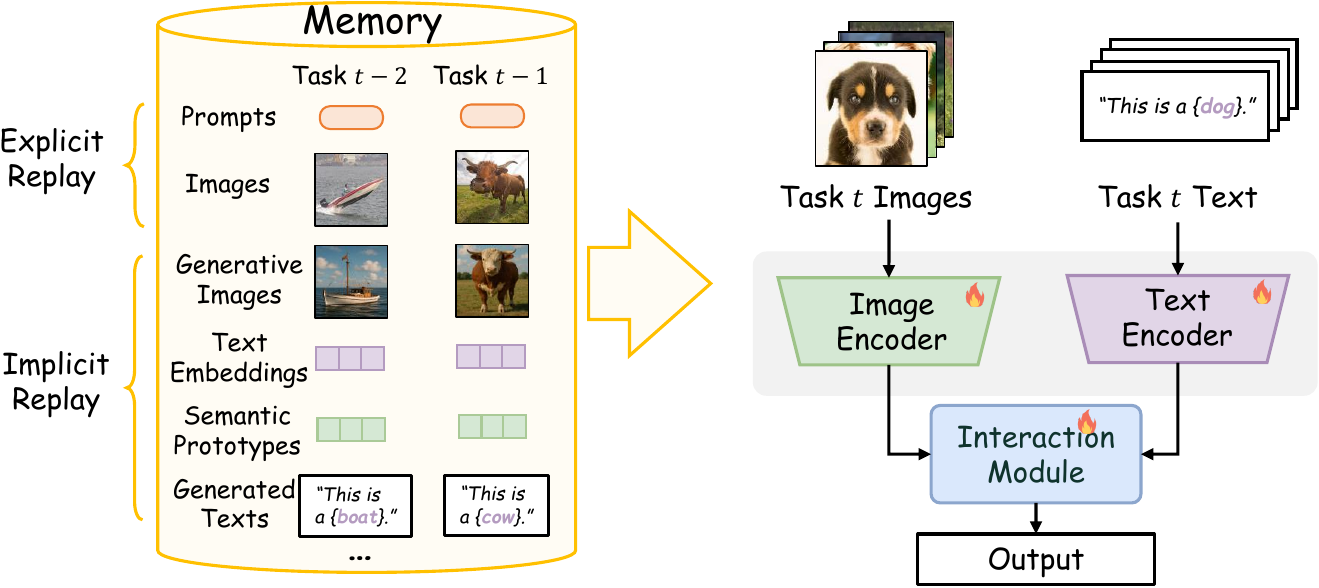} 
	\caption{\textbf{Schematic illustration of Multi-Modal Replay strategies.} To alleviate cross-modal feature drift and zero-shot capability erosion, MMRE methods operate at the data representation level. They can be divided into two paradigms: (1) \textbf{Explicit Replay} (top), which preserves previous knowledge through a memory buffer containing historical multimodal samples; and (2) \textbf{Implicit Replay} (bottom), which approximates previous data distributions using generative models or maintains semantic structures through prototype-based representations.}  
        \captionsetup{justification=centering}
	\label{fig:cls_mmre}  
\end{figure}

Replay strategies~\cite{rebuffi2017icarl,GEM,DGR} alleviate catastrophic forgetting by revisiting information from previous tasks and have become a fundamental approach in continual learning. In VLM-CL, replay is further utilized to address two key challenges: \textit{cross-modal feature drift} and \textit{zero-shot capability erosion}. By retaining and replaying information from earlier tasks, these methods help preserve modality alignment and maintain the general representation space learned during pre-training. However, directly replaying downstream task data may introduce overfitting due to limited memory diversity~\cite{huang2025MGCLIP}. Existing Multi-Modal Replay (MMRE) methods can be broadly divided into explicit and implicit replay strategies, as illustrated in Figure~\ref{fig:cls_mmre}.

\textbf{Explicit Replay} preserves previous knowledge by maintaining a subset of historical multimodal samples. Compared with naive rehearsal, which may lead to overfitting due to the limited diversity of memory samples, recent approaches focus on extracting more robust and transferable information from replay data. One direction explores \textit{feature disentanglement and representation stabilization}. For example, VQACL~\cite{zhang2023vqacl} separates sample representations into task-specific and task-invariant components, and stabilizes the invariant representations through prototype-based clustering to reduce semantic drift. Another direction investigates \textit{cross-modal interaction preservation}. MSPT~\cite{chen2023continual} uses the memory buffer to compute asymmetric distances and aligns important dimensions in cross-attention maps between historical and updated models. This strategy helps preserve the structure of modality fusion modules during continual adaptation.

% \textbf{Explicit Replay} directly stores and reuses a small subset of real data from previous tasks. For instance, VQACL~\cite{zhang2023vqacl} constructs a memory buffer and decouples sample features into specific and invariant components, updating the latter via prototypical clustering to maintain stability. Similarly, MSPT~\cite{chen2023continual} replays a small sample subset per task and uses an asymmetric distance function to align key dimensions in the attention maps of new and old models.

% \begin{figure}[!t] 
% 	\centering   
% 	\includegraphics[width = 1\linewidth]{figure/figure2.pdf} 
% 	\caption{Illustration of the detailed settings of different VLM continual learning methods. Methods are grouped into three taxonomies: (a) Multi-Modal Replay strategy, (b) Cross‑Modal Regularization strategy, and (c) Parameter‑Efficient Adaptation strategy, with the third category further divided into four subclasses.}  
%         \captionsetup{justification=centering}
% 	\label{fig:cls_details}  
% \end{figure} 

\textbf{Implicit Replay} reduces the storage and privacy limitations of explicit memory by approximating previous data distributions without retaining raw samples. Existing approaches mainly follow two directions: \textit{Generative Hallucination} and \textit{Prototype-driven Compression}. The first direction generates synthetic multimodal data to recover previous knowledge and maintain decision boundaries. For example, IncCLIP~\cite{yan2022generative} leverages frozen historical encoders to generate hard negative text embeddings, helping preserve task-specific boundaries during continual adaptation. More recent methods employ powerful generative models, such as diffusion models (e.g., Stable Diffusion~\cite{Rombach_2022_CVPR}), to synthesize visual exemplars from class-level semantic descriptions, as demonstrated in GIFT~\cite{wu2025synthetic}. In more complex VQA scenarios, implicit replay can also reconstruct intermediate semantic structures, including pseudo scene graphs (SGP~\cite{lei2023symbolic}) or pseudo-labels generated from previously stored textual questions (QUAD~\cite{marouf2025no}).

The second direction, \textit{Prototype-driven Compression}, avoids explicit generation of pixels or tokens and instead preserves the geometric structure of previously learned representations. Recent methods maintain visual and textual prototype memories, often updated through Exponential Moving Average (EMA), and combine them with current features to approximate historical contexts without storing raw data. For instance, MacVQA~\cite{li2026macvqa} dynamically maintains prototype pools for multimodal representation retention. Similarly, CGIL~\cite{frascaroli2024clip} preserves continuous class distributions through Variational Autoencoders (VAEs), enabling sampling from approximated historical distributions. To improve interpretability, language-guided Concept Bottleneck Models preserve knowledge through semantic prototypes~\cite{yu2025language}. Beyond memory representation, some approaches introduce architectural mechanisms to support implicit replay. AnytimeCL~\cite{zhu2024anytime} combines a dual-decoder architecture with a PCA-compressed memory bank, while CLL-CLIP~\cite{yang2024embracing} incorporates dynamically expandable text embedding layers to support continuous vocabulary expansion while maintaining previously learned representation structures.

% \textbf{Implicit Replay} simulates previous data distributions with generative models or pseudo-samples, mitigating storage and privacy concerns. For example, IncCLIP~\cite{yan2022generative} uses a previous model's encoder to generate hard negative text samples, which helps suppress feature drift by encouraging the model to distinguish between new and old tasks. In the VQA domain, SGP~\cite{lei2023symbolic} generates pseudo scene graphs, while QUAD~\cite{marouf2025no} takes a memory-efficient approach by storing only past questions to generate pseudo-labels for new images. 
% %new
% MacVQA~\cite{li2026macvqa} maintains separate EMA-updated pools of visual and textual prototypes and gate-fuses top-$k$ retrieved prototypes with the current feature.
% Other generative approaches include GIFT~\cite{wu2025synthetic}, which uses Stable Diffusion to synthesize images from class names, and CGIL~\cite{frascaroli2024clip}, which employs a variational autoencoder to learn class distributions for replay. Methods can also incorporate unique architectural elements; AnytimeCL~\cite{zhu2024anytime} uses a dual-decoder architecture with a PCA-compressed memory bank, while CLL-CLIP~\cite{yang2024embracing} features an expandable text embedding layer for continual language learning. To improve interpretability, Lu~\cite{yu2025language} applies language-guided Concept Bottleneck Models and replays knowledge using semantic prototypes.

In summary, multi-modal replay empirically mitigates zero-shot erosion and feature drift by operating at the data manifold level. Nevertheless, the inherent computational overhead, coupled with the persistent risk of overfitting to historical templates, propels the exploration of more mathematically constrained, parameter-level alternatives.

% In summary, the multi-modal replay strategy is a direct and effective approach for maintaining alignment and generalization by operating at the data level. However, its inherent storage and computational overhead has prompted researchers to explore more efficient alternatives.

\subsection{Cross-modal Regularization Strategy (CREG)}

\begin{figure}[!t] 
	\centering   
	\includegraphics[width = 0.9\linewidth]{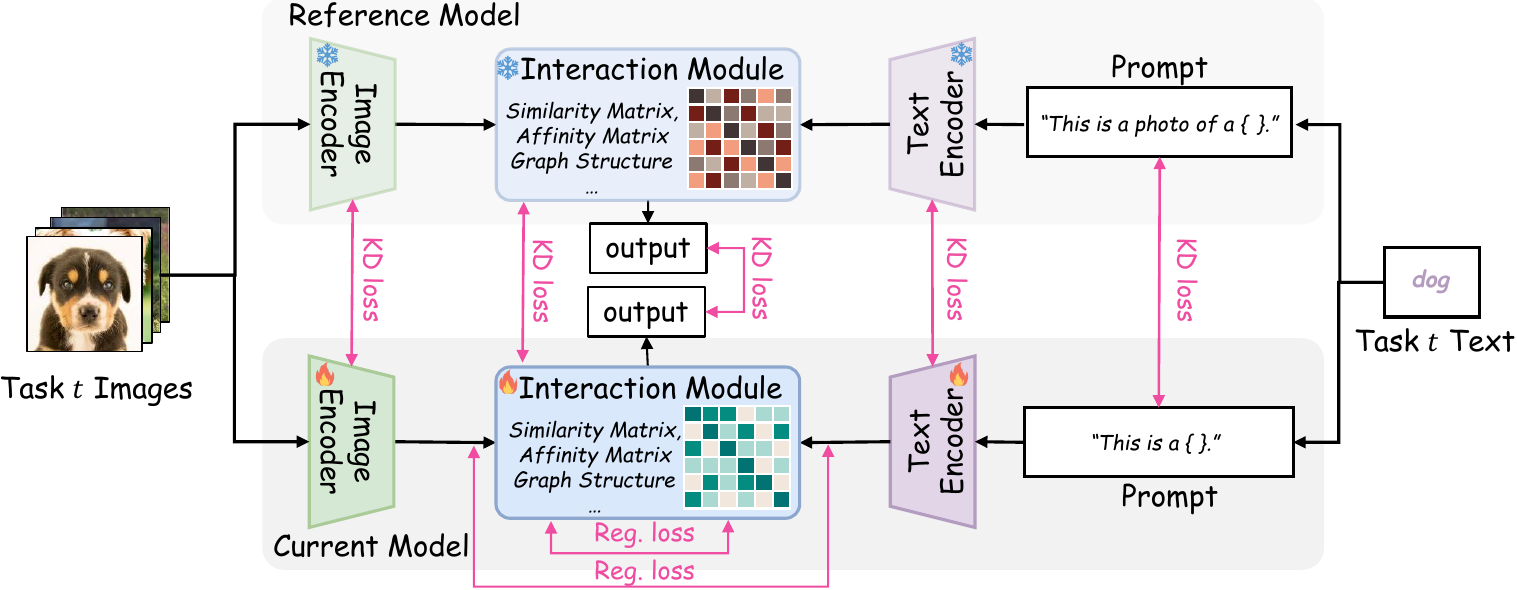} 
	\caption{\textbf{Architectural overview of Cross-Modal Regularization.} Unlike data-based replay approaches, CREG methods mitigate feature drift by introducing constraints during model optimization. By using a frozen reference model or historical model states as guidance, these methods preserve the structure of the pre-trained joint embedding space through regularization at different levels, including similarity distributions, affinity structures, and output representations.}  
        \captionsetup{justification=centering}
	\label{fig:cls_creg}  
\end{figure}

While Multi-Modal Replay provides a direct way to mitigate cross-modal feature drift, its dependence on stored or generated data introduces additional challenges, including memory consumption, privacy concerns, and potential overfitting to historical samples. Cross-Modal Regularization (CREG) addresses these limitations by avoiding explicit data rehearsal and instead preserving multimodal alignment through constraints imposed on the optimization process, as illustrated in Figure~\ref{fig:cls_creg}. Based on their underlying mechanisms, existing CREG methods can be grouped into four main paradigms.

\textbf{Cross-Modal Structure Preservation.} This paradigm focuses on maintaining the geometric structure of the pre-trained joint embedding space during continual adaptation. Instead of directly constraining model parameters, these methods preserve the relational relationships among multimodal representations. At the distribution level, ZSCL~\cite{zheng2023preventing} aligns the cross-modal similarity distributions of the current and initial models using a cross-entropy objective:
\begin{equation}
\mathcal{L}_{\text{dist\_img/text}} = \text{CE}(\mathbf{p}, \overline{\mathbf{p}}) = - \sum_{j=1}^{m} p_j \cdot \log \overline{p}_j,
\end{equation}
which helps maintain the original vision-language alignment. To preserve finer-grained semantic relationships, Mod-X~\cite{ni2023continual} and DKR~\cite{cui2024continual} constrain similarity structures or dynamically update vision-language affinity matrices. In the context of knowledge distillation, this idea extends beyond conventional logit matching toward structural preservation. For example, AwoForget~\cite{zheng2024adapt} constructs a cross-modal graph that captures both first-order image-text similarity and second-order intra-modal neighborhood relationships. SeGP-CL~\cite{he2026segpcl} further identifies drift-sensitive regions near task boundaries through adversarial anchors and applies targeted structural distillation. By preserving relative relationships and modality-specific separation, methods such as MG-CLIP~\cite{huang2025MGCLIP} help maintain zero-shot discrimination capability during continual adaptation.

\textbf{Orthogonal and Intrinsic Subspace Constraints.} To mathematically guarantee minimal interference with prior knowledge, a rigorous line of work projects new task updates into orthogonal or intrinsic low-energy parameter dimensions. This paradigm fundamentally restricts the optimization trajectory. For example, GNSP~\cite{peng2025gnspgradientnullspace} calculates the null space spanned by the activation features of all previously learned tasks and projects the current gradient exclusively into this orthogonal dimension. Parallelly, NuSA-CL~\cite{jo2025nusa} leverages Singular Value Decomposition (SVD) on weight matrices to identify and confine updates within task-agnostic intrinsic sub-spaces. Expanding beyond Euclidean geometry, HMCL~\cite{liu2026hyperbolic} enforces a shared hyperbolic isometry across modalities, projecting gradients to jointly preserve cross-modal relations and hierarchy-encoding radii. These strict mathematical constraints provide formidable stability, albeit necessitating careful calibration to prevent a severe degradation in network plasticity.

\textbf{Asymmetric Modality Regularization.} Recognizing that vision and text encoders degrade at profoundly different rates during downstream adaptation, this paradigm leverages the more stable modality (typically the invariant discrete logic of text) to anchor the highly volatile visual perception. DualTeacher~\cite{yu2024select} physically embodies this asymmetry by freezing the text encoder and exclusively updating the image encoder, regulating it via a weighted distillation from both pre-trained and historical models:
\begin{equation}
\mathcal{L}_{\text{KD}}^{\text{dual}} = \sum_{x \sim \mathcal{X}_{\text{ref}}} \eta(x) \cdot \mathcal{L}_{\text{KD}}^{t-1} + (1 - \eta(x)) \cdot \mathcal{L}_{\text{KD}}^{0}.
\end{equation}
Beyond standard distillation, textual semantics actively direct the optimization of visual features. Both RAPF~\cite{huang2024class} and SECA~\cite{he2025seca} heavily exploit text-embedding affinity: RAPF enforces a large-margin separation (hinge loss) between new visual features and related old text categories, while SECA utilizes this textual similarity to dynamically weight instance-adaptive distillation and refine visual prototypes. To further suppress spurious correlations, where the model simply relies on language shortcuts, SCD~\cite{lao2023multi} introduces counterfactuals (e.g., zero-padded images) into the asymmetric distillation pipeline. Conversely, visual cues can also anchor text components; CPE-CLIP~\cite{d2023multimodal} and CLAP4CLIP~\cite{jha2024clap4clip} deploy visual prompts and visually-guided attention to prevent text feature collapse.

\textbf{Fusion-Level and Hybrid Regularization.} For VLMs with deep cross-modal fusion architectures, regularization can be applied directly to cross-attention modules. CTP~\cite{zhu2023ctp} employs symmetric cross-entropy and masking strategies to balance the contribution of different modalities, while MSPT~\cite{chen2023continual} and TRIPLET~\cite{qian2023decouple} regulate attention updates through adaptive gradient scaling or low-rank cross-modal prompt constraints. Beyond module-level constraints, architectural approaches such as PROOF~\cite{zhou2025learning} introduce constrained task-specific projection layers to reduce representation drift. In practice, combining multiple regularization mechanisms can provide stronger protection against forgetting. For example, GIFT~\cite{wu2025synthetic} integrates knowledge distillation, Fisher information-based constraints, and synthetic data alignment to jointly regulate parameter updates.

Overall, cross-modal regularization avoids the storage requirements of explicit replay by incorporating alignment constraints into the optimization process. Instead of directly preserving model parameters, these methods focus on maintaining representation structures through approaches such as topology preservation, orthogonal projection, and asymmetric guidance. However, their effectiveness depends on a careful understanding of VLM architectures and appropriate selection of regularization strength to balance stability and plasticity during continual adaptation.
 
% \textbf{Guided Regularization} uses one modality to inform the learning of another. For instance, RAPF~\cite{huang2024class} uses text similarity to identify related old categories and applies a hinge loss to enforce visual feature separation from new classes.
% %new
% SECA~\cite{he2025seca} uses CLIP's textual semantics to both weight an instance-adaptive distillation from a pool of historical adapters and refine visual-side prototypes via inter-class text-embedding affinity.
% %感觉可以两者合一起
% %For instance, RAPF~\cite{huang2024class} and SECA~\cite{he2025seca} both leverage textual similarity to semantically related old categories as guidance for visual learning: RAPF separates new features from them via a hinge loss, while SECA distills from them and refines visual prototypes via their affinity.
% Similarly, CPE-CLIP~\cite{d2023multimodal} and CLAP4CLIP~\cite{jha2024clap4clip} use visual prompts and visual-guided attention, respectively, to ensure text features remain aligned with visual cues. GIFT~\cite{wu2025synthetic} is a composite strategy, combining distillation, alignment with a matrix from synthetic data, and Fisher information constraints to regulate weight updates.

% In summary, cross-modal regularization offers data-efficient strategies to maintain VLM alignment during continual learning. These methods effectively mitigate feature drift via soft constraints, avoiding the storage overhead of replay. However, their success often depends on a nuanced understanding of VLM mechanisms and careful tuning of regularization strength to balance stability and plasticity.

\subsection{Parameter-Efficient Adaptation Strategy (PEA)}

\begin{figure}[!t] 
	\centering   
	\includegraphics[width = 1\linewidth]{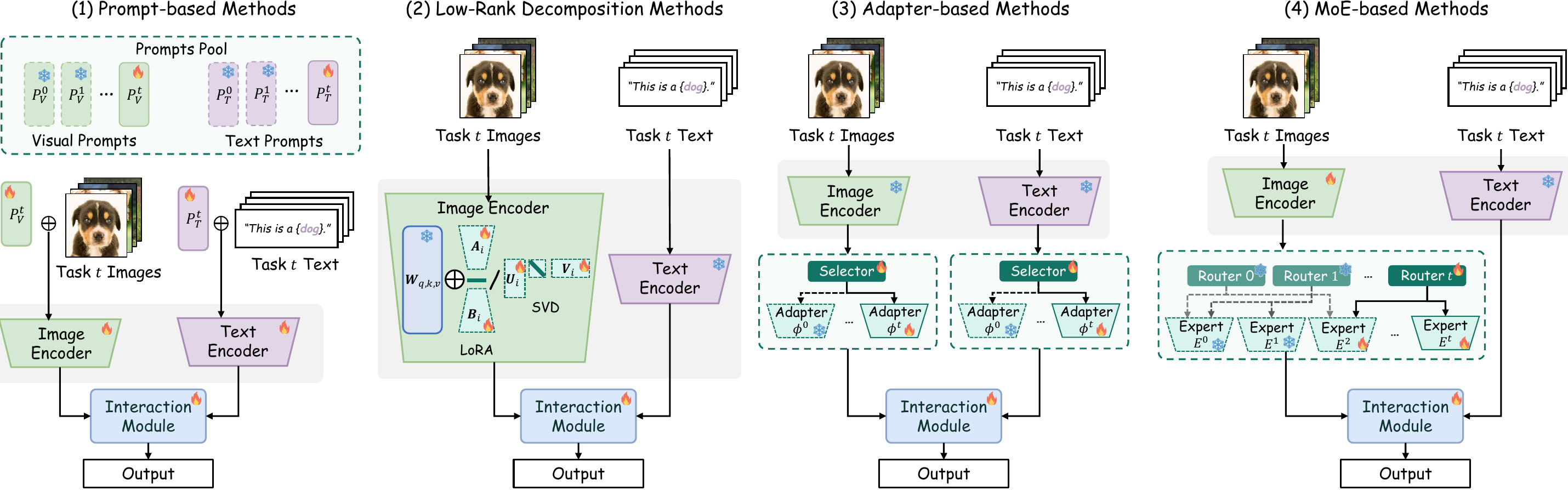} 
	\caption{\textbf{Intrinsic mechanisms of Parameter-Efficient Adaptation (PEA).} To reduce shared module interference, PEA methods isolate task-specific updates while keeping the majority of pre-trained parameters frozen. The taxonomy covers four representative paradigms: (1) prompt-based methods that introduce task-specific semantic representations at the input level, (2) low-rank decomposition methods that allocate task-specific updates within compact subspaces, (3) adapter-based methods that insert lightweight modules for task adaptation, and (4) Mixture-of-Experts (MoE) methods that enable capacity sharing through adaptive routing. Blue modules denote frozen pre-trained parameters, while red modules represent trainable task-specific components.}  
        \captionsetup{justification=centering}
	\label{fig:cls_pea}  
\end{figure}

While cross-modal regularization provides an effective way to constrain representation drift, applying regularization across the full parameter space of large foundation models can further restrict model plasticity during continual adaptation. In particular, excessive constraints on shared components, such as cross-attention modules, may limit the ability to acquire new concepts. Parameter-Efficient Adaptation (PEA) addresses this limitation by restricting updates to a small set of task-specific parameters or modular components while keeping the majority of the pre-trained backbone fixed. This design reduces interference among tasks and helps preserve zero-shot capabilities by avoiding extensive modifications to shared representations~\cite{biderman2024lora}. Rather than categorizing methods solely by their insertion locations (e.g., input embeddings or attention layers), we organize PEA approaches according to four underlying mechanisms related to parameter allocation and task isolation, as illustrated in Figure~\ref{fig:cls_pea}.

% Despite the elegance of cross-modal regularization in mathematically constraining feature drift, applying soft penalties across the entire, massive parameter space of foundation models inevitably exacerbates the stability-plasticity dilemma. Regularizing shared cross-attention mechanisms often severely limits the model's plasticity for acquiring novel downstream concepts. To overcome this limitation of global optimization, Parameter-Efficient Adaptation (PEA) provides an architectural paradigm shift: it enforces strict structural isolation by freezing the core backbone and restricting all gradient updates to localized, modular sub-networks. This paradigm addresses the core challenge of \textit{shared module interference} by freezing the main pre-trained model and updating only a small number of new or designated parameters. This approach structurally isolates updates for each task, preventing catastrophic overwrites of shared weights. By preserving the core model, PEA methods also effectively alleviate \textit{zero-shot capability erosion}, a benefit supported by prior findings~\cite{biderman2024lora}. We categorize these methods into four main types, as shown in Figure~\ref{fig:cls_details} (c).

\textbf{Prompt-Based Parameter Isolation.}
Prompt-based methods adapt models by introducing learnable prompt vectors while keeping the backbone parameters fixed. A key challenge is how to balance task-specific adaptation and knowledge sharing across tasks. Early approaches mainly adopted independent prompt learning for different tasks. For example, S-liPrompt~\cite{wang2022s} learns separate prompts for each task and freezes previous prompts after training, improving stability but limiting knowledge transfer.

Recent studies have explored more flexible prompt designs by decomposing prompts into shared and task-specific components. LPI~\cite{yan2024low} factorizes a three-dimensional prompt tensor $\mathbf{P} \in \mathbb{R}^{i \times j \times k}$ into low-rank factors:
\begin{equation}
    \mathbf{P}[i][j][k] = \mathrm{AVG}\left( \mathbf{D_1}[i] \odot \mathbf{D_2}[j] \odot \mathbf{D_3}[k] \right),
\end{equation}
where $\mathbf{D_1}$ captures shared information, while $\mathbf{D_2}$ and $\mathbf{D_3}$ encode modality-specific and task-related factors. Similarly, PromptCCZSL~\cite{maryam2025promptcczsl} separates prompts into session-aware compositional representations and session-independent attribute representations. Other approaches regulate prompt updates through gradient scaling (CPE-CLIP~\cite{d2023multimodal}) or multi-teacher distillation~\cite{maryam2025promptcczsl} to preserve previous knowledge while enabling adaptation to new tasks.

\textbf{Low-Rank Adaptation and Subspace Constraints.}
Low-rank adaptation methods, such as LoRA~\cite{hu2022lora}, restrict parameter updates to low-dimensional subspaces and have become widely used for continual adaptation. However, sharing the same low-rank space across tasks can still introduce interference. Existing approaches mainly address this issue through subspace constraints and adaptive capacity allocation.

For subspace constraints, LW2G~\cite{feng2024lw2g} and KeepLoRA~\cite{luo2026keeplora} reduce interference by projecting new task updates onto directions orthogonal to previously learned subspaces or historical feature representations. Alternatively, adaptive capacity allocation methods adjust the contribution of different low-rank components. CoDyRA~\cite{lu2024adaptive} learns a weighted combination of factorized matrices through SVD-based decomposition:
\begin{equation}
    \Delta \mathbf{W} = \sum_{i=1}^r \alpha_i \mathbf{B}_i \mathbf{A}_i,
\end{equation}
where $\boldsymbol{\alpha}$ controls the importance of different rank components. DAC~\cite{fa2026dac} further decomposes LoRA modules into orthogonal rank-1 experts and uses token-guided sparse composition for adaptive parameter selection. Liu et al.~\cite{liu2026branch} explore the low-rank parameter space through gradient-free exploration to improve adaptation flexibility.

\textbf{Adapter-Based Parameter Consolidation.}
Adapter-based methods introduce lightweight trainable modules, typically implemented as bottleneck networks, to achieve parameter-efficient adaptation. However, independently maintaining adapters for different tasks increases parameter costs and may introduce inconsistencies between task-specific representations.

Recent studies therefore focus on adapter consolidation and stabilization. C-CLIP~\cite{liu2025c} dynamically merges task-specific adapters into the backbone, while CLAP4CLIP~\cite{jha2024clap4clip} adopts a probabilistic consolidation strategy. To reduce interference between isolated modules, ATLAS~\cite{li2024atlas} and DIKI~\cite{tang2024mind} introduce orthogonal constraints and zero-initialized residual branches. Beyond Euclidean representations, HypCL~\cite{cheng2026hypcl} models task-specific adapters in the Poincaré ball and applies Möbius addition to compose sequential updates. Other approaches use stable references, such as textual embeddings~\cite{zhang2025visual} or channel masks~\cite{zhang2024learning}, to guide adapter optimization.

\textbf{MoE-Based Adaptive Routing.}
Mixture-of-Experts (MoE) architectures~\cite{jacobs1991moe, noam2017sg-moe} provide another direction for parameter-efficient adaptation by dynamically selecting task-relevant experts. Given a routing function $\mathcal{R}^t$ and expert set $\mathcal{E}_i$, MoE-Adapters~\cite{yu2024boosting} compute representations as:
\begin{equation}
    y^t = \sum_i \gamma_i^t \cdot \mathcal{E}_i(x^t).
\end{equation}

Although MoE improves parameter utilization, routing stability remains a major challenge. MoE-Confidence~\cite{forhad2026mixing} regularizes newly learned routers using the predictive confidence of previous experts. Recent studies also suggest that routing behaviors differ across modalities. Visual representations are often easier to separate due to their spatial structure, whereas language representations are more context-dependent, making routing decisions more sensitive to instruction variations.

\textbf{Limitations of Parameter-Efficient Adaptation.}
Despite reducing parameter interference, PEA methods remain constrained by limited adaptation capacity. As the number of tasks increases, fixed prompt pools, low-rank spaces, and predefined expert sets may become insufficient to accommodate additional knowledge. Furthermore, strict orthogonality constraints used in methods such as LW2G and ATLAS may reduce available optimization directions, potentially limiting model plasticity and knowledge transfer.

Overall, PEA provides an important approach for VLM-CL by controlling updates through selective parameter adaptation. However, individual mechanisms based on prompts, low-rank updates, adapters, and MoE routing each face scalability challenges. Recent studies increasingly explore hybrid strategies that combine PEA with complementary approaches, such as cross-modal regularization and representation replay~\cite{yu2024exploiting, zhang2024overcoming}, to achieve a better balance between stability, plasticity, and computational efficiency.

%% file: methods_mllms.tex
\section{Extending the VLM-CL Framework to MLLMs}\label{sec:method_mllms}

The continual learning strategies discussed in Section~\ref{sec:method}, including Multi-Modal Replay, Cross-Modal Regularization, and Parameter-Efficient Adaptation, were initially developed to preserve the stability of joint representation spaces in predictive VLMs. With the rapid emergence of generative Multimodal Large Language Models (MLLMs)~\cite{liu2023visual,bai2023qwen,alayrac2022flamingo}, the focus of continual learning has gradually shifted from representation preservation toward maintaining the broader capabilities of generative models.

This transition does not invalidate existing CL paradigms; instead, it requires adapting their objectives to the characteristics of autoregressive generation. Although the visual perception module continues to experience feature drift, the integration of an LLM backbone introduces additional challenges associated with next-token prediction, instruction following, and complex reasoning. Therefore, MLLM continual learning aims not only to preserve multimodal representations, but also to maintain generative behaviors, reasoning trajectories, and alignment properties acquired during instruction tuning.

Recent studies have demonstrated that MLLMs remain susceptible to catastrophic forgetting, while conventional CL strategies exhibit varying effectiveness in this generative setting~\cite{he2026continual_inst_mllm}. For example, replay-based approaches and model expansion strategies often provide stronger protection against forgetting compared with direct regularization. A growing body of research has investigated continual learning for representative MLLMs, including LLaVA~\cite{liu2023visual}, Qwen2.5-VL~\cite{bai2023qwen,bai2025qwen25vl}, MiniGPT-v2~\cite{chen2310minigpt}, and InternVL2.5~\cite{chen2025expanding}, as summarized in Figure~\ref{fig:cls-method}. Recent developments further explore hierarchical task-specific expansion~\cite{guo2025hide}, LoRA/MoE-based continual instruction tuning~\cite{yu2025progressive,ge2025dynamic}, continual VQA~\cite{huai2025clmoe}, and federated continual instruction tuning~\cite{guo2025federated}. The following sections discuss the emerging challenges of MLLM-CL and examine how existing CL paradigms have been adapted to address these issues.

\subsection{Emergent Challenges in MLLM-CL} \label{subsec:mllm_challenges}

Compared with conventional VLMs, where forgetting is mainly reflected in the distortion of multimodal representations~\cite{zhu2023ctp, zheng2023preventing}, generative MLLMs introduce additional challenges due to their autoregressive generation process and heterogeneous model components. Recent studies have identified several emerging issues in MLLM continual learning, particularly in maintaining instruction alignment, reasoning ability, and cross-modal interaction.

\textbf{Instruction Alignment and Behavioral Forgetting.}
MLLMs are typically aligned with human preferences and interaction formats through instruction tuning or reinforcement learning from human feedback (RLHF). However, continual fine-tuning on narrow downstream domains may introduce an ``alignment tax''~\cite{liu2025llavac}, causing the model to gradually lose previously acquired alignment behaviors. Such degradation is not limited to factual knowledge forgetting, but may also affect response formats, safety-related behaviors, and instruction-following ability~\cite{alssum2025unforgotten}. Consequently, models may retain certain domain-specific knowledge while becoming less consistent in following user instructions or generating structured responses~\cite{chen2024coinbenchmarkcontinualinstruction, chen2025sefe}.

\textbf{Reasoning Capability Degradation.}
Preserving complex reasoning ability is another important challenge in MLLM continual learning~\cite{gao2026many, guo2026mllmctbench}. Unlike conventional CL settings that mainly focus on retaining task-related knowledge, MLLMs must preserve multi-step reasoning processes encoded in the LLM backbone. Continual updates may alter internal representations associated with spatial, logical, and compositional reasoning, leading to inconsistent intermediate reasoning steps or reduced ability to infer conclusions from visual evidence. Such degradation can occur even when basic recognition and factual knowledge remain relatively stable.

\textbf{Cross-Modal Alignment and Modality Imbalance.}
Most MLLMs adopt an asymmetric architecture consisting of a vision encoder, a modality projector, and a large language model backbone~\cite{liu2023visual,li2023blip2}. During continual adaptation, optimization updates may favor linguistic representations over visual features, causing the model to rely more heavily on textual priors while reducing the contribution of visual information~\cite{deng2025words,li2026easier}. This phenomenon, often referred to as visual blindness, reflects a degradation of cross-modal interaction during generation. Meanwhile, excessive updates to the modality projector may disturb the mapping between visual representations and the LLM embedding space, weakening the multimodal alignment learned during pre-training~\cite{gao2026many}.

Although Multi-Modal Replay (MMRE) remains applicable to MLLM-CL in principle, its practical use is constrained by high computational costs and potential distribution shifts introduced by replayed conversational data. Therefore, recent studies have increasingly explored cross-modal regularization (Section~\ref{subsec:mmlm-creg}) and parameter-efficient adaptation (Section~\ref{subsec:mmlm-pea}) as more scalable approaches for preserving multimodal alignment and generative capabilities.

\subsection{CREG Methods for MLLM-CL} \label{subsec:mmlm-creg}

The cross-modal regularization mechanisms introduced in Section~\ref{sec:method} can be naturally extended to MLLM continual learning. In this setting, regularization is no longer limited to constraining representation drift, but also aims to preserve generative policies, preference alignment, and reasoning behaviors.

$\phi$-DPO~\cite{truong2026phidpo} extends the idea of policy regularization by reformulating the KL constraint with respect to previous policies into a Direct Preference Optimization (DPO)~\cite{rafailov2023direct} objective. By incorporating reweighting strategies, it improves robustness under imbalanced multimodal continual learning scenarios.

CGL~\cite{yao2026cgl} investigates continual GUI learning by balancing supervised fine-tuning (SFT) and group-relative policy optimization (GRPO)~\cite{shao2024deepseekmath} through an entropy-guided scheduling strategy. It further projects SFT gradients onto GRPO anchor directions to reduce conflicting optimization components.

Beyond representation-level constraints, CPO++~\cite{yang2026robust} targets reasoning drift during multimodal reinforcement fine-tuning. By combining counterfactual reasoning and perception-level perturbations with preference optimization, it aims to reduce spurious correlations and improve the robustness of multimodal reasoning.

\subsection{PEA Methods for MLLM-CL}\label{subsec:mmlm-pea}

Although regularization-based approaches such as DPO can constrain policy drift during continual adaptation, updating all parameters of large MLLM backbones remains computationally expensive and memory-intensive. Therefore, Parameter-Efficient Adaptation (PEA) has become an increasingly popular strategy for continual visual instruction tuning, with LoRA-based methods and their extensions receiving particular attention due to their favorable trade-off between adaptation capacity and computational cost.

LoRA-based continual tuning has demonstrated strong effectiveness in MLLM-CL. Li~\textit{et al.}~\cite{li2026easier} show that a simple adaptation scheme, consisting of incremental task-specific LoRA modules, a conservative learning rate, and a small amount of mixed general-purpose instruction data, can achieve competitive performance compared with more complex continual learning pipelines. Building upon this observation, subsequent studies have explored how to improve the scalability, knowledge sharing, and task management ability of LoRA-based adaptation.

One line of research focuses on progressive adapter expansion and capacity management. Progressive LoRA~\cite{yu2025progressive} maintains a task-indexed LoRA pool and combines task-aware allocation with task recall mechanisms to reuse previous knowledge and recover alignment with earlier tasks. D-MoLE~\cite{ge2025dynamic} further introduces dynamic layer-wise LoRA expert allocation under a constrained parameter budget, together with a gradient-based inter-modal curriculum to alleviate task interference and modality imbalance. PCLR~\cite{meng2026pclr} decomposes LoRA modules into rank-level experts and adopts a compression--integration--learning procedure to recycle adapter capacity, reducing forgetting while controlling model expansion.

Another direction investigates structured parameter sharing and selective updating to improve stability during continual adaptation. SEFE~\cite{chen2025sefe} distinguishes between superficial and essential forgetting, addressing them through answer-style diversification and LoRA regularization, respectively. BranchLoRA~\cite{zhang2025enhancing} selectively updates or freezes different branches of the model, enabling task specialization while preserving cross-task cooperation. LiLoRA~\cite{che2026lilora} introduces a shared representation mechanism by maintaining a common down-projection $\mathbf{A}$ across tasks and decomposing each up-projection into a shared basis $\mathbf{B}_0$ with task-specific low-rank residuals $\tilde{\mathbf{B}}_i\tilde{\mathbf{A}}_i$. Cosine regularization is further applied to reduce drift in the shared basis.

Beyond manually assigned adapters, routing-based approaches provide a more adaptive mechanism for selecting task-relevant parameters. MR-LoRA~\cite{zhao2025mllmclcontinuallearningmultimodal} separates task- and domain-specific LoRA experts and employs an adapted MLLM as the routing module. SMoLoRA~\cite{wang2025smolora} further separates visual understanding and instruction-following adapters to preserve different functional capabilities. HiDe-LLaVA~\cite{guo2025hide} introduces a hierarchical MoE framework to organize progressively accumulated LoRA modules, while CL-MoE~\cite{huai2025clmoe} combines task-level and instance-level routing with dynamic momentum updates for continual adaptation. SAME~\cite{xie2026same} constrains router optimization within spectral subspaces, incorporates historical covariance information for expert updates, and selectively freezes previously important experts. Hyper-LLaVA~\cite{xu2026hyperllava} models task distributions in the Poincaré ball space and adjusts visual-textual contributions according to routing uncertainty.

In addition to LoRA-based solutions, several studies explore alternative parameter-efficient mechanisms. MAny~\cite{gao2026many} extends training-free consolidation to multimodal continual instruction tuning through prototype-guided projector merging and recursive least-squares LoRA fusion, aiming to preserve both perception and reasoning capabilities. ModalPrompt~\cite{zeng2025modalprompt} stores task-specific knowledge using prototype prompts and performs cross-modal prompt selection and fusion. Instruction-Grounded Visual Projectors~\cite{jin2025instruction} employ instruction-conditioned mixtures of visual projectors with expert recommendation and pruning strategies, providing another approach for balancing visual representation learning and instruction following.

%% file: Evaluation.tex
\section{Evaluation and Performance Analysis}\label{sec:evaluation}

Evaluating continual learning (CL) in vision-language models (VLMs) necessitates a multifaceted approach that addresses both traditional CL desiderata (\textit{e.g.}, mitigating catastrophic forgetting, balancing stability-plasticity) and VLM-specific challenges (\textit{e.g.}, preserving cross-modal alignment, zero-shot generalization). This section synthesizes established evaluation protocols, specialized metrics, and benchmark datasets tailored to capture the unique complexities of VLM-based CL. 

\begin{figure}[!t]  
	\centering 
	\includegraphics[width = 1\linewidth]{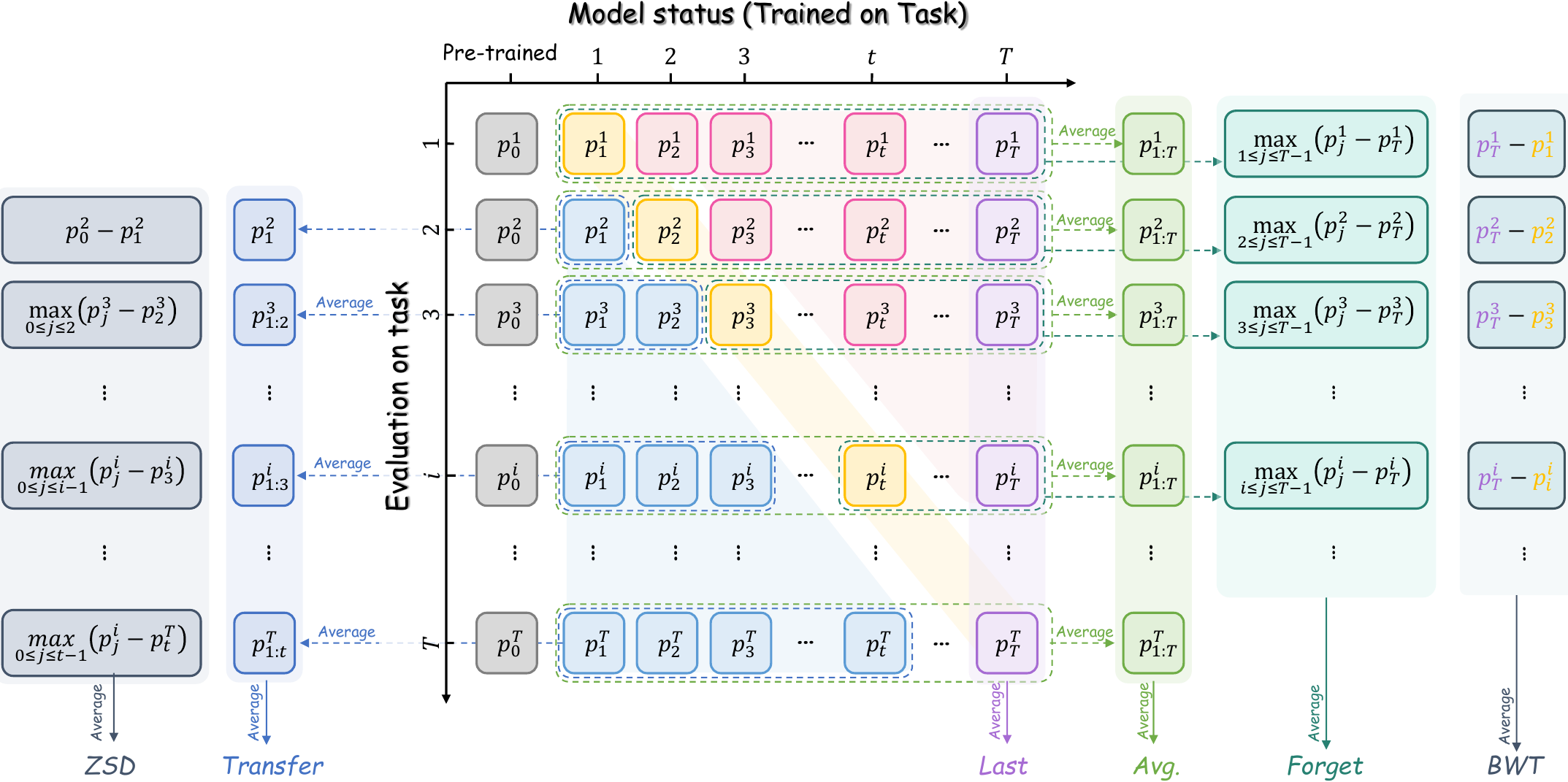}  
	\caption{Schematic illustration of the metrics commonly employed in VLM‑CL. The matrix shows performance from its pretrained state through to after completing training on the final task (horizontal axis) and evaluating on all tasks (vertical axis). Colored regions represent task‐wise accuracy at different stages, alongside the corresponding processed metric values.} 
        \captionsetup{justification=centering}
	\label{Fig_metric} 
\end{figure}

\subsection{Core Evaluation Metrics}
The assessment of VLM-CL methods relies on a hierarchy of metrics designed to quantify performance across sequential tasks, generalization capabilities, and resource efficiency. To improve readability we assign a color to each metric, as shown in Figure~\ref{Fig_metric}. 

\textcolor{avg}{\textbf{Average Accuracy}}~\cite{yan2022generative,d2023multimodal,yu2024boosting,park2024pre,li2024atlas,zheng2023preventing,cao2024generative} remains fundamental, measuring mean task performance over all learning stages to gauge overall proficiency. 
\begin{equation}
    \textcolor{avg}{\textit{\textbf{Avg.}}}=\frac{1}{T} \sum_{t=1}^{T} \left(\frac{1}{T} \sum_{i=1}^{T} p_i^{(t)}\right).
\end{equation}

Complementarily, \textcolor{last}{\textbf{Last Accuracy}}~\cite{liu2023class,yu2024boosting,zheng2023preventing,xu2024advancing,tang2024mind,park2024pre,zhou2025learning} evaluates the model’s retained competence after full training, reflecting practical deployability.
\begin{equation}
    \textbf{\textit{\textcolor{last}{Last}}} = \frac{1}{T} \sum_{t=1}^{T}p_T^{(t)}.
\end{equation}

To diagnose knowledge retention, \textcolor{forget}{\textbf{Forgetting Ratio}}~\cite{lao2023multi,yu2024select,qian2023decouple,marouf2025no,cui2024continual,yang2024embracing}  quantifies the maximum performance drop per task post-initial learning.
\begin{equation}
    \textbf{\textit{\textcolor{forget}{Forget}}}=\frac{1}{T-1} \sum_{t=1}^{T-1} \max_{t \leq i \leq T}\left( p_i^{(t)} - p_t^{(t)} \right).
\end{equation}

Meanwhile, \textcolor{bwt}{\textbf{Backward Transfer}}~\cite{jha2024clap4clip,lei2023symbolic} assesses improvements or regressions on earlier tasks induced by later learning.
\begin{equation}
    \textbf{\textit{\textcolor{bwt}{BWT}}}=\frac{1}{T-1} \sum_{t=1}^{T-1} \left(p_T^{(t)} -p_t^{(t)}  \right).
\end{equation}

Crucially, VLMs demand metrics beyond task-specific adaptation. \textcolor{transfer}{\textbf{Zero-shot Transfer}}~\cite{yu2024boosting,zheng2023preventing,xu2024advancing,tang2024mind,wu2025synthetic} evaluates generalization to unseen tasks using pre-acquired knowledge.
\begin{equation}
    \textbf{\textit{\textcolor{transfer}{Transfer}}} =\frac{1}{T-1} \sum_{t=2}^{T} \left(\frac{1}{t-1} \sum_{i=1}^{t-1} p_i^{(t)}\right).
\end{equation}

\textcolor{zsd}{\textbf{Zero-Shot Degradation}}~\cite{yu2024select} explicitly measures erosion of this capability, a critical vulnerability in VLMs.
\begin{equation}
    \textcolor{zsd}{\textbf{\textit{ZSD}}}=\frac{1}{T-1} \sum_{t=2}^{T} \max_{1 \leq i \leq t-1}\left( p_1^{(1)} - p_i^{(t)} \right).
\end{equation}

For retrieval tasks, \textbf{Recall$@$K}~\cite{ni2023continual,yan2024low,cui2024continual,zhou2025learning,yang2024embracing} and \textbf{mean Average Precision}~\cite{zhu2023ctp, zhang2024learning} are indispensable. The definition formulas for these two metrics are as follows:
\begin{align}
    R@K&=\frac{|\mathcal{R}_q\cap\{ d_1, d_2,\ldots,d_K\} |}{|\mathcal{R}_q|}\\
    mAP&= \frac{1}{Q} \sum_{i=1}^{Q} \frac{1}{m_q} \sum_{k=1}^{K} P_q(k) \delta_q(k), 
\end{align}
where $P_q(k) = \frac{\mathcal{R}_q(k) + 1}{k}$. These are further decomposed into \textbf{Image-to-Text} (\textit{I2T}) and \textbf{Text-to-Image} (\textit{T2I)} variants to diagnose asymmetric forgetting. VQA-centric benchmarks adopt \textbf{Answer Precision} (\textit{AP})~\cite{schutze2008introduction} and task-specific \textit{F1} scores~\cite{blair1979information}.

\input{table/benchmark}

\subsection{Evolution of Benchmarks for VLM-CL and MLLM-CL}
The landscape of benchmarks for vision-language continual learning can be categorized into three evolutionary tiers based on their design specificity and evaluation focus (Table~\ref{tab:vlm_cl_benchmark}).

\colorbox{runi}{\textbf{Repurposed unimodal benchmarks}} adapt computer vision datasets for CL evaluation by reducing text modalities to categorical labels, offering scalability but limited cross-modal insight. TinyImageNet~\cite{chrabaszcz2017downsampled}, CIFAR-100~\cite{krizhevsky2009learning}, ImageNet-100~\cite{deng2009imagenet}, and CLEAR-10/100~\cite{lin2021clear} test robustness to class/domain shifts. DomainNet~\cite{peng2019moment} (6-domain DIL) evaluates domain robustness but compresses text to class names. CUB-200~\cite{wah2011caltech} serves as small-scale CIL testbeds. While valuable for scalability testing, these benchmarks cannot diagnose asymmetric modality degradation or compositional failure. Conceptual 12M~\cite{changpinyo2021conceptual} extends CL to dense prediction but lack compositional language interaction. CDDB~\cite{li2023continual} and CORe50~\cite{lomonaco2017core50} specialize in forensic and robotic applications but ignore textual reasoning. Although these benchmarks are useful for evaluating scalability and adaptation efficiency, they provide limited insights into VLM-specific forgetting mechanisms. For example, performance degradation on CIFAR-100 mainly reflects visual recognition forgetting, while DomainNet-based DIL cannot distinguish whether degradation originates from visual representations or cross-modal alignment changes.

\colorbox{amul}{\textbf{Adapted multimodal benchmarks}} reconfigure existing datasets for CL evaluation while retaining cross-modal interactions. MDL-VQA~\cite{lao2023multi} sequences diverse visual domains (natural, art, medical) in domain-incremental learning (DIL), exposing modality reliability shifts when transitioning from high-resolution natural images to noisy medical scans. The P9D~\cite{zhu2023ctp} industrial retrieval benchmark (1M+ samples) combines long-tailed category distributions with incremental product additions, challenging models to maintain fine-grained alignment under real-world data imbalances. These benchmarks preserve multimodal interactions while introducing continual learning protocols, providing a closer approximation to real-world vision-language adaptation scenarios.

\colorbox{native}{\textbf{VLM/MLLM-Specific Benchmarks}} explicitly address cross-modal forgetting pathologies through innovative task formulations. For compositional reasoning assessment, VQACL~\cite{zhang2023vqacl} introduces dual evaluation regimes—standard testing and novel, composition testing, where disentangled visual concepts and logical operations are recombined to quantify zero-shot generalization decay. In retrieval domains, COCO-CL~\cite{perez2020incremental} extends the classic MSCOCO~\cite{lin2014microsoft} with sequential task splits while monitoring cross-modal retrieval stability alongside traditional detection metrics.
For continual pretraining scenarios, temporal web-scale benchmarks like TIC-CC3M/TIC-YFCC~\cite{garg2023tic} (12M-15M samples) provide chronological splits of web-crawled image-text data, directly evaluating zero-shot capability erosion through time-decayed Recall$@$1. 
For generative MLLMs, emerging instruction-tuning benchmarks (CoIN~\cite{chen2024coinbenchmarkcontinualinstruction}, UCIT~\cite{guo2025hide}, MLLM-CL~\cite{zhao2025mllmclcontinuallearningmultimodal}, CVIT~\cite{wang2025smolora}) evaluate continual adaptation across diverse tasks, including visual question answering, reasoning, and OCR-related scenarios. These benchmarks further consider instruction-following consistency and capability retention beyond conventional accuracy-based evaluation.
MLLM-CTBench~\cite{guo2026mllmctbench} further introduces process-level reasoning diagnosis, whereas CLeaRS~\cite{clears} extends evaluation to remote-sensing scenarios. ToS~\cite{kong2026eca} further organizes image-to-text streams by dominant visual topics, preserving cross-task semantic overlap to model realistic distribution shifts.

%Current VLM-CL evaluation relies predominantly on class-incremental protocols (73\% of Table~\ref{tab:vlm_cl_benchmark} datasets), with native multimodal benchmarks (30\%) providing critical but sparse coverage of zero-shot decay and compositional robustness. Adapted benchmarks (45\%) offer scaled diversity yet obscure modality interactions, while repurposed unimodal datasets (25\%) prioritize efficiency over cross-modal diagnostics. This current fragmentation highlights the necessity of establishing more coherent and unified protocols.

Table~\ref{tab:vlm_cl_benchmark} reveals a shift from repurposed unimodal datasets and adapted benchmarks toward protocols designed specifically for VLM-CL and MLLM-CL. Although class-incremental evaluation remains prevalent, the current landscape is fragmented: repurposed datasets favor scalability and efficiency but offer limited cross-modal diagnostics, adapted benchmarks provide greater diversity but may obscure modality interactions, and native benchmarks better capture zero-shot decay, compositional transfer, instruction following, domain or ability acquisition, and unseen-task generalization. This observation motivates future benchmark designs that combine standard continual learning metrics with modality-aware and capability-oriented evaluations.

%% file: table/benchmark.tex
\begin{table*}[!t]
\centering
\renewcommand{\arraystretch}{1.4}
\tiny
% \scriptsize
% \footnotesize
\caption{Benchmark for VLM Continual Learning. Different colors (\colorbox{runi}{\rule{0pt}{0.1cm}\rule{0.1cm}{0pt}}, \colorbox{amul}{\rule{0pt}{0.1cm}\rule{0.1cm}{0pt}}, \colorbox{native}{\rule{0pt}{0.1cm}\rule{0.1cm}{0pt}}) indicate repurposed Unimodal, adapted multimodal and native benchmark for VLM-CL, respectively.}
\label{tab:vlm_cl_benchmark}
% Use the NiceTabular environment instead of tabular
\begin{NiceTabular}{m{35pt} m{52pt} m{38pt} m{30pt} m{52pt} m{33pt} m{43pt}}
\toprule
\textbf{Category} & \textbf{Dataset} & \textbf{Task Type} & \textbf{Scenario} & \textbf{Modality} & \textbf{Size} & \textbf{Metrics} \\
\midrule

% --- Repurposed Unimodal Block ---
% \Block replaces \multirow and handles coloring itself. No more \rowcolor needed!
\Block[fill=runi]{10-1}{\makecell[c]{Repurposed\\Unimodal}} & CDDB\cite{li2023continual} & Classification & DIL & Image & \~50K & Accuracy \\ 
& CORe50\cite{lomonaco2017core50} & Classification & DIL & Video & 50K & Accuracy \\ 
& DomainNet\cite{peng2019moment} & Classification & DIL & Image & 600K & Accuracy \\
& Conceptual12M\cite{changpinyo2021conceptual} & Segmentation & CIL / IIL & Image+Text & 12M & IoU, AP \\
& ImageNet-100/1K\cite{deng2009imagenet} & Classification & CIL / DIL & Image & 130K–1.3M & Accuracy \\
& TinyImageNet\cite{chrabaszcz2017downsampled} & Classification & CIL & Image & 100K & Accuracy \\
& CIFAR100\cite{krizhevsky2009learning} & Classification & CIL & Image & 60K & Accuracy \\
& CUB200\cite{wah2011caltech} & Classification & CIL & Image & 11.7K & Accuracy \\
& CLEAR-10/100\cite{lin2021clear} & Classification & CIL / DIL & Image & 4.3M–18.6M& Accuracy \\
& ADE20K-CL\cite{zhou2017scene} & Segmentation & CIL & Image+Annotations & 25K & IoU \\
\midrule\midrule

% --- Adapted Multimodal Block ---
\Block[fill=amul]{5-1}{\makecell[c]{Adapted\\Multimodal}}  & MDL-VQA\cite{lao2023multi} & VQA & DIL & Image+Text & 150K & Accuracy \\
& P9D\cite{zhu2023ctp} & Retrieval & DIL / TIL & Image+Text & 1M+ & AP \\
& Flickr30K\cite{young2014image} & Retrieval & CIL & Image+Annotations & 30K & Accuracy \\
& ECommerce-T2I\cite{M6} & Retrieval & CIL & Image+Annotations & 15K & Accuracy \\
& NExT-QA\cite{xiao2021next} & VQA & TIL & Videos+Text & 52K & AP \\
\midrule\midrule

% --- Native VLM-CL Block (no color) ---
% Here we use \Block without the fill option, so it just acts like a better \multirow
\Block[fill=native]{16-1}{\makecell[c]{VLM/MLLM-\\Specific \\Benchmarks}} & CliMB\cite{srinivasan2022climb} & VQA & TIL & Image+Text & 1.1M & Accuracy, F1 \\
& VQACL\cite{zhang2023vqacl} & VQA & TIL / DIL & Image+Text & \~100K & Accuracy \\
& COCO-CL\cite{perez2020incremental} & Seg. / Ret. & CIL & Image+Annotations & 200K+ & AP, IoU, Acc. \\
& TiC\cite{garg2023tic} & Ret. / Class. & Time-IL & Image+Text & 127M/1B/12B & Acc., Recall \\
& CoIN\cite{chen2024coinbenchmarkcontinualinstruction} & VQA & TIL & Image+Text & 740K & Accuracy \\
& UCIT\cite{guo2025hide} & VQA & TIL & Image+Text & 231K & Accuracy \\
& MLLM-CL\cite{zhao2025mllmclcontinuallearningmultimodal} & VQA & DIL / AIL & Image+Text & 392.7K/925K& Accuracy \\
& MLLM-CTBench\cite{guo2026mllmctbench} & VQA & TIL & Image+Text & 70K & Accuracy \\
& CLeaRS\cite{clears} & VQA & DIL / TIL & Image+Text & 207K & Accuracy \\
& ToS\cite{kong2026eca} & VQA/Cap. & DIL & Image+Text & 935K & Accuracy \\
& MTIL\cite{zheng2023preventing} & Classification & TIL / CIL & Image & 438.3K & Accuracy \\
& VTAB\cite{zhai2019large} & Classification & CIL & Image & 10K & Accuracy \\
& CLOVE\cite{lei2023symbolic} & VQA & DIL / TIL & Image+Text & N/A & Accuracy \\
& OMNI\cite{li2026omnibench} & Classification & CIL & Image & 1M+ & Accuracy \\
& IMRE\cite{chen2023continual} & RE & TIL & Image+Text & 9K & F1 \\
& IMNER\cite{chen2023continual} & NER & TIL & Image+Text & 8.5K & F1 \\
\bottomrule
\end{NiceTabular}
\end{table*}

%% file: experiments.tex
\subsection{Performance Analysis on Key Tasks}
By leveraging the above metrics and benchmarks, we can analyze the performance of existing SOTA methods in common VLM-CL application scenarios.

\subsubsection{Image Classification}
Image classification is a classic task in computer vision and is also widely applied in multimodal settings. VLM-based continual learning methods particularly focus on leveraging multimodal interactions to enhance visual classification capability throughout the continual learning process. In normal settings, datasets are typically introduced sequentially under the paradigms of Class-Incremental Learning (CIL), Domain-Incremental Learning (DIL), or Task-Incremental Learning (TIL). Task-Id must be provided both the training and inference stages in TIL specifically. 

In this scenario, classification accuracy is typically employed alongside various conventional continual learning metrics to evaluate the performance of VLM-based continual learning methods. 

\input{table/classification_result}

Talbe \ref{tab:class_acc} shows the accuracy of several SOTA methods on benchmark MTIL with metrics \textcolor{transfer}{\textit{Transfer}}, \textcolor{avg}{\textit{Average}} and \textcolor{last}{\textit{Last}}. 
Unseen classes are used to evaluate the model's \textbf{zero-shot performance}. While downstream tasks may weaken the generalization ability of the pretrained model, some methods freeze the pretrained model and rely on its original zero‑shot capabilities~\cite{lu2024boosting}. Alternatively, parameter‑efficient techniques inject PEFT modules into the backbone after each task, leveraging knowledge transfer to bolster zero‑shot accuracy~\cite{liu2025c}.
\textbf{Few-shot performance} reflects a model’s capacity to rapidly adapt to a new task using only a handful of training examples. To this end, multi-modal replay-based approaches enrich the scarce training set with rehearsal samples drawn from previous tasks. Regularization‑based techniques impose soft constraints on parameter updates to stabilize learning under data scarcity~\cite{yu2024exploiting}. Parameter‑efficient modules further boost few‑shot adaptability by limiting the number of tunable parameters, reducing overfitting risk while preserving core representations~\cite{zhang2025visual}.
Some approaches employ \textbf{long-sequence tasks} to evaluate the extreme limits of incremental learning capabilities, which aligns with the real-world scenarios of continual learning. Multi-Modal replay-based methods are often limited by rapidly growing memory requirements. However some methods compress the rehearsal unit to tackle this challenge~\cite{jha2024clap4clip}\cite{cao2024generative}. Parameter‑efficient schemes inject lightweight modules that are repeatedly reused and integrated across tasks, enabling scaling to longer sequences without a proportional increase in storage~\cite{feng2024lw2g}\cite{wang2023attriclip}.

\subsubsection{Multimodal Retrieval}
Retrieval tasks in VLM-CL involve searching for images given text queries (\textit{T2I}) or vice-versa (\textit{I2T}) from a continuously growing data pool. This requires maintaining fine-grained local and global feature alignment. As shown in Table \ref{tab:retrival_acc}, on standard benchmarks like Flickr30K~\cite{young2014image} and MS-COCO~\cite{lin2014microsoft}, continual fine-tuning without a proper CL strategy leads to a significant performance drop compared to joint training. Methods like DKR~\cite{cui2024continual}, which performs dynamic knowledge rectification, and Mod-X~\cite{ni2023continual}, which preserves off-diagonal similarity information, demonstrate strong performance, nearly matching the zero-shot CLIP baseline and significantly outperforming traditional methods like EWC. This highlights the importance of strategies that actively preserve the relational structure of the multimodal embedding space.

\input{table/retrival_result}

\subsubsection{Visual Question Answering (VQA)}

Visual Question Answering (VQA) requires models to jointly understand visual content and question semantics. In continual settings, it evaluates the retention of previously learned visual--language mappings and the acquisition of new reasoning skills. 
As shown in Table \ref{tab:vlm_cl_benchmark}, VQA-CL benchmarks have emerged in recent years. 
CLiMB~\cite{srinivasan2022climb} constructs continual task streams from multiple VQA datasets, whereas VQACL~\cite{zhang2023vqacl} additionally evaluates novel compositions of reasoning skills and visual concepts. 
With the advent of continual instruction tuning for MLLMs, corresponding benchmarks have been developed concurrently. Leveraging more capable base models, these new benchmarks encompass a broader array of diverse and complex tasks, as reflected by the native benchmarks summarized in Table~\ref{tab:vlm_cl_benchmark}.

\begin{figure}[t!] 
  \centering
  \includegraphics[width=\linewidth]{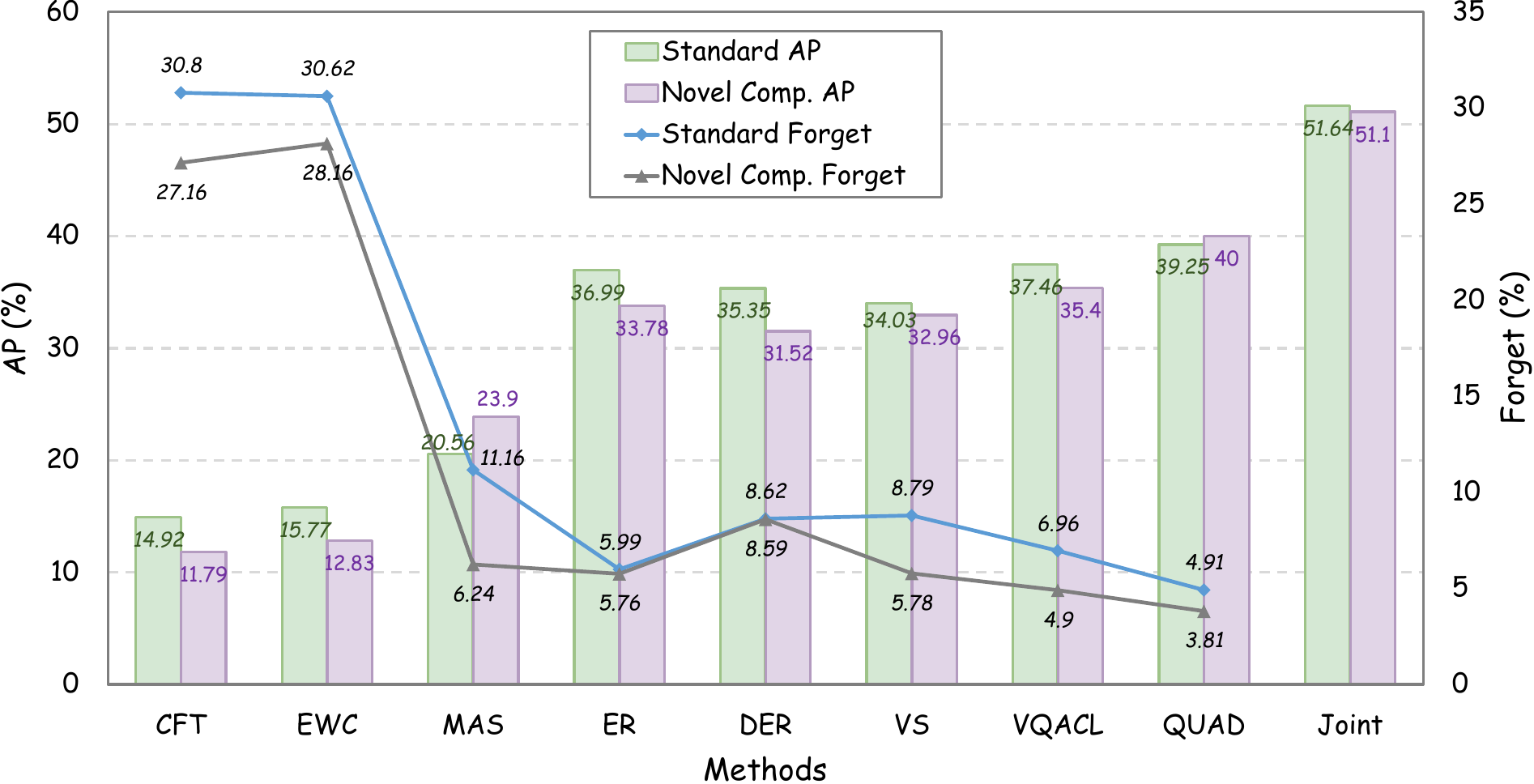}
  \caption{Comparison of SOTA methods on VQACL benchmark, where data extracted from \cite{marouf2025no}. It evaluates \textbf{\textit{AP}} and \textcolor{forget}{\textit{\textbf{Forget}}} on both the original data and novel compositions.}
  \label{fig:vqa_result}
\end{figure}

Figure \ref{fig:vqa_result} presents the baseline performance of SOTA on VQA benchmarks. Among them, QUAD \cite{marouf2025no} achieves outstanding results owing to its combination of pseudo‑label memory and attention distillation. Under similar configurations, TRIPLET \cite{qian2023decouple}, which utilizes decoupled prompts to capture complex relationships between modalities, and SCD \cite{lao2023multi}, which proposes a self-critical distillation mechanism to enforce consistency in the cross‑modal semantics, demonstrate distinct advantages in cross-modal interaction.

\subsubsection{Other Scenarios}
In addition to the scenarios mentioned above, VLM-CL approaches can be evaluated in various other settings or mixture of them, reflecting their broad applicability in real-world applications. TAM~\cite{zhang2024learning} applies matrix fusion to class‑incremental object detection, achieving strong performance on both COCO 2017~\cite{lin2014microsoft}and Pascal‑VOC 2007~\cite{pascal-voc-2007} splits.  MSPT~\cite{chen2023continual} proposes two new benchmarks, IMRE and IMNER, for relation extraction and named entity recognition tasks, further addressing cross‑modal challenges by modulating learning rates per modality. Apart from retrieval tasks, LPI~\cite{yan2024low} is also evaluated on referring expression comprehension tasks on benchmark RefCOCO~\cite{kazemzadeh2014referitgame}, where it localizes targets within complex visual scenes with minimal additional parameters. CLL‑CLIP~\cite{yang2024embracing} constructs a multilingual VLM benchmark over MSCOCO and XM3600 (36 languages)~\cite{thapliyal2022crossmodal} to evaluate incremental image‑assisted language retrieval. It expands CLIP via a token embedding layer and cross‑modal objectives, showing substantial gains without catastrophic forgetting.

\input{table/method_cost}

\subsubsection{Parameter and Memory Efficiency}
A crucial aspect of evaluation, especially for parameter-efficient adaptation strategies, is the computational cost. Table~\ref{tab:cost} shows a comparison of the number of trainable and additional parameters for various methods extracted from \cite{lu2024adaptive}. Multi-modal replay-based methods often incur extra memory costs for storing exemplars. Among these, implicit replay approaches generally require less additional storage than explicit replay, owing to their tailored compression strategies as discussed in \cite{marouf2025no}. In contrast, PEFT methods like MoE-Adapters~\cite{yu2024boosting} and CoDyRA~\cite{lu2024adaptive} dramatically reduce the number of trainable parameters compared to full fine-tuning, demonstrating their efficiency and practicality for real-world deployment where computational resources are limited. The additional parameter overhead for cross‑modal regularization varies with the choice of reference module, such as network‑based~\cite{lao2023multi} or embedding‑based~\cite{yu2024exploiting}, while for parameter‑efficient adaptation methods (e.g.,ATLAS~\cite{li2024atlas} and C-CLIP~\cite{liu2025c}), it scales with both the architecture of the adaptation module and the increase strategy. This efficiency is a key advantage that drives the adoption of such methods in VLM-CL, especially important for MLLM-CL with large-scale backbones.

%% file: table/classification_result.tex
\begin{table}[t]
\centering
\caption{Comparison of SOTA methods on the MTIL benchmark. Performance metrics are extracted directly from their respective original publications.}
\label{tab:class_acc}
\begin{tabular*}{0.98\columnwidth}{l@{\extracolsep{\fill}}ccc}
\toprule
\textbf{Method}       & \textcolor{transfer}{\textit{\textbf{Transfer}}} & \textcolor{avg}{\textit{\textbf{Avg.}}} & \textcolor{last}{\textit{\textbf{Last}}} \\
\midrule\midrule
\textbf{Zero-shot}              & 69.4             & 65.3          & 65.3          \\
\textbf{Continual Finetune}      & 44.6             & 55.9          & 77.3          \\
\textbf{Full Finetune}            & N/A             & 89.2          & N/A          \\
\midrule\midrule
LwF~\cite{li2017learning}       & 56.9             & 64.7          & 74.6          \\
iCaRL~\cite{rebuffi2017icarl}     & 50.4             & 65.7          & 80.1          \\
LwF-VR~\cite{ding2022don}    & 57.2             & 65.1          & 76.6          \\
WiSE-FT~\cite{wortsman2022robust}   & 52.3             & 60.7          & 77.7          \\
ZSCL~\cite{zheng2023preventing}      & 68.1             & 75.4          & 83.6          \\
L2P~\cite{wang2022learning}      & 53.2             & 67.9          & 82.0          \\
DualPrompt~\cite{wang2022dualprompt}    & 52.4             & 68.0          & 82.3          \\
S-liPrompt~\cite{wang2022s}      & 52.2             & 68.3          & 83.4          \\
MoE-Adapter~\cite{yu2024boosting} & 68.9           & 76.7          & 85.0          \\
Dual-RAIL~\cite{xu2024advancing}   & 69.4           & \textbf{77.8}         & 86.8          \\
AwoForget~\cite{zheng2024adapt}   & \textbf{69.8}        & 76.9          & 85.1          \\
DIKI~\cite{tang2024mind}      & 67.4             & 75.7          & 85.1          \\
DPeCLIP~\cite{lu2024boosting}      & 69.1            & 77.5          & \textbf{86.9}          \\
GIFT~\cite{wu2025synthetic} & 69.3             & 77.3          & 86.0          \\
\bottomrule
\end{tabular*}
\end{table}

%% file: table/retrival_result.tex
% \usepackage{multirow}
\begin{table}[t]
\caption{Comparison of SOTA methods on retrieval benchmarks. It records \textit{I2T} and \textit{T2I} R@1 of different models, where data is extracted from \cite{cui2024continual}.}
\label{tab:retrival_acc}
 \begin{tabularx}{\columnwidth}{Xcc cc cc}
\toprule
        \multirow{2}{*}{\textbf{Methods}} & \multicolumn{2}{c}{\textbf{Flickr30K}} & \multicolumn{2}{c}{\textbf{MS-COCO}} & \multicolumn{2}{c}{\textbf{EC}} \\ 
        \cmidrule(lr){2-3} \cmidrule(lr){4-5} \cmidrule(lr){6-7}
        & \textit{I2T} & \textit{T2I} & \textit{I2T} & \textit{T2I} & \textit{I2T} & \textit{T2I} \\ 
        \midrule\midrule
        \textbf{Joint}              & 64.5  &  46.9 & 39.8    & 22.2 &  23.5   & 23.5 \\ 
        \textbf{Zero-shot}          & 77.7   & 58.9   & 50.1  & 30.2  & 11.3  & 10.1   \\
        \textbf{CFT}                & 63.4   & 44.4   & 36.8  & 20.6   & 16.6   & 15.8   \\\midrule\midrule
        EWC\cite{kirkpatrick2017overcoming}   & 64.0   & 44.8  & 37.7    & 20.7  & 16.2   &  16.5 \\
        Mod-X\cite{ni2023continual}         & 73.1    &  55.6 & 47.1   & 27.9  & 20.1    & 20.0  \\
        DKR\cite{cui2024continual}       & \textbf{78.5}    &\textbf{58.7}  &\textbf{51.7}   &\textbf{29.7}  &  \textbf{20.4} & \textbf{20.2}  \\ 
\bottomrule
\end{tabularx}
\end{table}

%% file: table/method_cost.tex
\begin{table}[t]
\centering
\caption{Comparison of computational cost extracted from \cite{lu2024adaptive}. It records the count of training parameters and additional parameters/memory of some VLM-based methods training on MTIL benchmark under the experiment setting.}
\label{tab:cost}
\begin{tabular*}{\columnwidth}{l@{\extracolsep{\fill}}cc}
\toprule
\textbf{Methods} & \textbf{Training} & \textbf{Additional} \\
\midrule\midrule
LWF\cite{li2017learning}         & 129.6M            & None               \\
ZSCL\cite{zheng2023preventing}        & 129.6M            & None               \\
MoE-Adapters\cite{yu2024boosting} & 59.8M            & 13.35M             \\
RAIL\cite{xu2024advancing}        & \textit{N/A}      & 24.18M / 9.01M     \\
CoDyRA\cite{lu2024adaptive}           & 4.4M             & None               \\
\bottomrule
\end{tabular*}
\end{table}

%% file: future_work.tex
\section{Discussion \& Future-Work}\label{sec:discussion}
This survey has systematically reviewed the landscape of continual learning for vision-language models, charting its unique challenges, dominant methodologies, and evaluation frameworks. Our analysis reveals that VLM-CL is maturing from a simple extension of unimodal techniques into a distinct field with its own fundamental problems. We now discuss the broader implications of our findings and outline promising directions for future research.

\subsection{Discussion}
Our proposed challenge-driven taxonomy, Multi-Modal Replay (MMRE), Cross-Modal Regularization (CREG), and Parameter-Efficient Adaptation (PEA), goes beyond superficial architectural differences to deconstruct the intrinsic mechanisms of multimodal forgetting. Across these paradigms, we observe a definitive mechanistic evolution: from explicit raw data rehearsal to geometric manifold simulation, from absolute parameter freezing to topology-preserving orthogonal projections, and from naive structural insertion to dynamic subspace allocation.

Crucially, our synthesis reveals that the universally acknowledged stability-plasticity dilemma manifests asymmetrically across modalities in VLMs. Visual representations often exhibit spatially clusterable patterns conducive to dynamic routing and localized updates, whereas highly entangled language instructions frequently trigger representation collapse under sequential shifting. Furthermore, we must critically acknowledge the theoretical ``capacity ceiling.'' Relying strictly on parameter isolation or rigorous orthogonal projection eventually consumes all available degrees of freedom in high-dimensional spaces as the task sequence extends infinitely ($T \to \infty$), culminating in a severe degradation of network plasticity. Consequently, isolated strategies are no longer mathematically sufficient. The field is witnessing an irreversible pivot toward hybrid frameworks that systematically couple parameter-efficient structural isolation with cross-modal topological regularization.

Finally, as our review demonstrates, the vast majority of current research operates safely within the Continual Fine-Tuning (CFT) paradigm. While leveraging frozen foundation models offers a highly efficient, pragmatic solution for constrained downstream applications, this static backbone assumption fundamentally restricts the acquisition of novel, open-world semantics. As the frontier expands rapidly from predictive cross-modal alignment toward the autoregressive generation of Multimodal Large Language Models (MLLMs) and the continuous physical action spaces of Embodied Agents, the optimization objective fundamentally shifts. The ultimate trajectory of VLM-CL must therefore transcend static CFT, evolving toward dynamic Continual Pre-Training (CPT) and hierarchical architectures capable of unbounded capacity expansion and robust logical reasoning preservation.

\subsection{Future Work}
Based on the gaps and trends identified in this survey, we propose several key directions for future research:
\paragraph{Unified and Holistic Benchmarking} The current evaluation landscape is fragmented (see Table~\ref{tab:vlm_cl_benchmark}). Future benchmarks should move beyond simple classification accuracy and incorporate metrics that specifically diagnose the core challenges we identified. We advocate for a new generation of benchmarks that mandate: (1) Compositional zero-shot evaluation to explicitly measure ZSD by testing novel combinations of known concepts. (2) Disentangled modality metrics to quantify cross-modal feature drift, for instance, by reporting separate performance scores for the vision and text encoders on unimodal tasks. (3) Temporal, web-scale data streams, following the lead of datasets like TiC, to properly evaluate models in realistic, lifelong CPT scenarios.

%\paragraph{Advancing Continual Pre-training (CPT)} To make CPT viable, research should focus on ``scalable forgetting mitigation.'' This could involve developing importance weighting methods (like EWC~\cite{kirkpatrick2017overcoming}) that are efficient enough for billion-parameter models, or exploring new forms of dynamic architectures and replay. For instance, can a model learn to generate its own ``pseudo-replays'' from its internal knowledge to consolidate learning without storing real data? Answering these questions is critical for creating foundation models that do not become obsolete.
\paragraph{Continual Learning for Generative and Interactive Tasks}Most current VLM-CL research focuses on discriminative tasks like classification and retrieval. The next frontier lies in endowing generative and interactive models with continual learning capabilities. How can a multimodal chatbot continually learn from its conversations with a user without forgetting past interactions? How can a robot continually learn to follow new types of instructions in an ever-changing environment? These scenarios involve complex challenges like learning from feedback, managing long-term memory, and ensuring safety and alignment over time.

\paragraph{Continual Learning for Vision-Language-Action (VLA) Models}
Transitioning from static tasks to Embodied AI, Vision-Language-Action (VLA) models (e.g., RT-2~\cite{zitkovich2023rt}, OpenVLA~\cite{kim2024openvla}) map cross-modal inputs to continuous, high-frequency action spaces. This fundamentally alters catastrophic forgetting: cross-modal feature drift no longer merely degrades accuracy but precipitates cascading physical failures, where misaligned semantics cause embodied agents to fail safety-critical instructions~\cite{yoo2024exploratory}. To mitigate this, future architectures must adopt hierarchical decoupling and skill-compositional strategies~\cite{zhang2026learning, jia2025hierarchical}. Such designs allow upper-layer visual semantics to adapt continuously to non-stationary environments while rigorously preserving the stability of low-level control policies (e.g., via reusable skill experts or incremental task-aware routing). Consequently, static metrics (e.g., mAP) are inadequate; the field requires dynamic evaluation frameworks that quantify sequential success rates and cumulative physical safety under interactive environmental feedback.

\paragraph{Towards a Theoretical Understanding} The field is currently driven by empirical success. A deeper theoretical understanding of VLM-CL is needed. Can we mathematically model cross-modal feature drift? Can we derive theoretical bounds on how much new information a PEA module can learn before it interferes with pre-trained knowledge? A stronger theoretical foundation would enable us to move from creating ad-hoc solutions to designing principled, provably effective continual learning algorithms for VLMs. 
Furthermore, future works could explore how the modality gap affects continual learning in VLMs. Modality gap~\cite{liang2022mind} refers to the gap between the text and image embeddings in the shared VLM embedding space. While recent works~\cite{schrodi2025two,mistretta2025cross,levidouble} extensively analyze the modality gap in VLMs, this has not been discussed in the context of continual learning.